\documentclass[11pt]{article}

\usepackage[final]{acl}

\usepackage{times}
\usepackage{latexsym}

\usepackage[T1]{fontenc}

\usepackage[utf8]{inputenc}

\usepackage{microtype}

\usepackage{inconsolata}

\usepackage{graphicx}
\usepackage{amsmath}
\usepackage{amssymb}
\usepackage{multirow}
\usepackage{booktabs}
\usepackage{listings}
\usepackage{xcolor}
\definecolor{MyDarkOrange}{RGB}{166, 82, 0}

\title{CARL: Constraint-Aware Reinforcement Learning for Planning with LLMs}

\author{
    Qiuyi Qi$^\spadesuit$$^\diamondsuit$\footnotemark[1],
    Jinjian Zhang$^\diamondsuit$\thanks{Q. Qi, J. Zhang and M. Bao contributed equally to this work.},
    Mutian Bao$^\spadesuit$$^\diamondsuit$\footnotemark[1], 
    Tian Liang$^\spadesuit$$^\diamondsuit$,
    Guocong Li$^\spadesuit$$^\diamondsuit$,\\ 
    \textbf{Dongnan Liu}$^\diamondsuit$,
    \textbf{Wei Zhou}$^\diamondsuit$,
    \textbf{Jie Liu}$^\clubsuit$,
    \textbf{Ming Kong}$^\spadesuit$\footnotemark[2],\\
    \textbf{Linjian Mo}$^\diamondsuit$\footnotemark[2],
    \textbf{Feng Zhang}$^\spadesuit$,
    \textbf{Qiang Zhu}$^\spadesuit$\thanks{Q. Zhu, M. Kong and L. Mo are corresponding authors. Q. Zhu is with the College of Artificial Intelligence, Shanghai Institute for Advanced Study, Zhejiang University. M. Kong is with the School of Earth Sciences, Zhejiang University. L. Mo is with the Ant Group.}\\
    $^\spadesuit$ Zhejiang University, 
    $^\diamondsuit$ Ant Group,
    $^\clubsuit$ City University of Hong Kong \\
    \texttt{
    \{qiqiuyi,zhuq\}@zju.edu.cn 
    }
}

\begin{document}
\maketitle
\begin{abstract}
Despite their strong reasoning capabilities and extensive world knowledge, Large Language Models (LLMs) frequently generate plans that violate task constraints, undermining their reliability in real-world applications. This deficiency arises from a lack of systematic mechanisms to incorporate constraint information during the generation process. While existing approaches attempt to mitigate this by relying on external tools or task decomposition, they fail to enhance the model's intrinsic constraint awareness. To address this, we propose Constraint-Aware Reinforcement Learning (CARL), a novel RL framework designed to strengthen LLMs' intrinsic focus on constraints. 
CARL introduces a constraint-aware reward by comparing the model's output distributions under constrained and unconstrained inputs, encouraging constraint focus and penalizing neglect.
Compatible with various RL frameworks and requiring no external solvers or top models, CARL enables scalable, end-to-end constraint-aware planning. 
Extensive experiments on BlocksWorld, TravelPlanner, and T-Eval demonstrate that CARL significantly outperforms standard Reinforcement Fine-Tuning (RFT) baselines and state-of-the-art reasoning models, exhibiting a markedly increased focus on constraints.
\end{abstract}

\section{Introduction}
\label{sec:intro}

Large Language Models (LLMs) have demonstrated remarkable capabilities in reasoning, tool utilization, and world knowledge modeling, positioning them as powerful candidates for complex planning tasks—a cornerstone of cognitive AI systems~\citep{huang2022inner,ahn2022can}. Planning entails generating a sequence of executable actions to achieve a goal while strictly adhering to a set of constraints~\citep{newell1958elements,kartam1990towards}. For instance, a travel itinerary must not only satisfy the destination and timeline but also comply with specific constraints such as \textit{budget limits}, \textit{transportation preferences}, or \textit{dietary requirements}.

\begin{figure}[t]
\centering
  \includegraphics[width=\columnwidth]{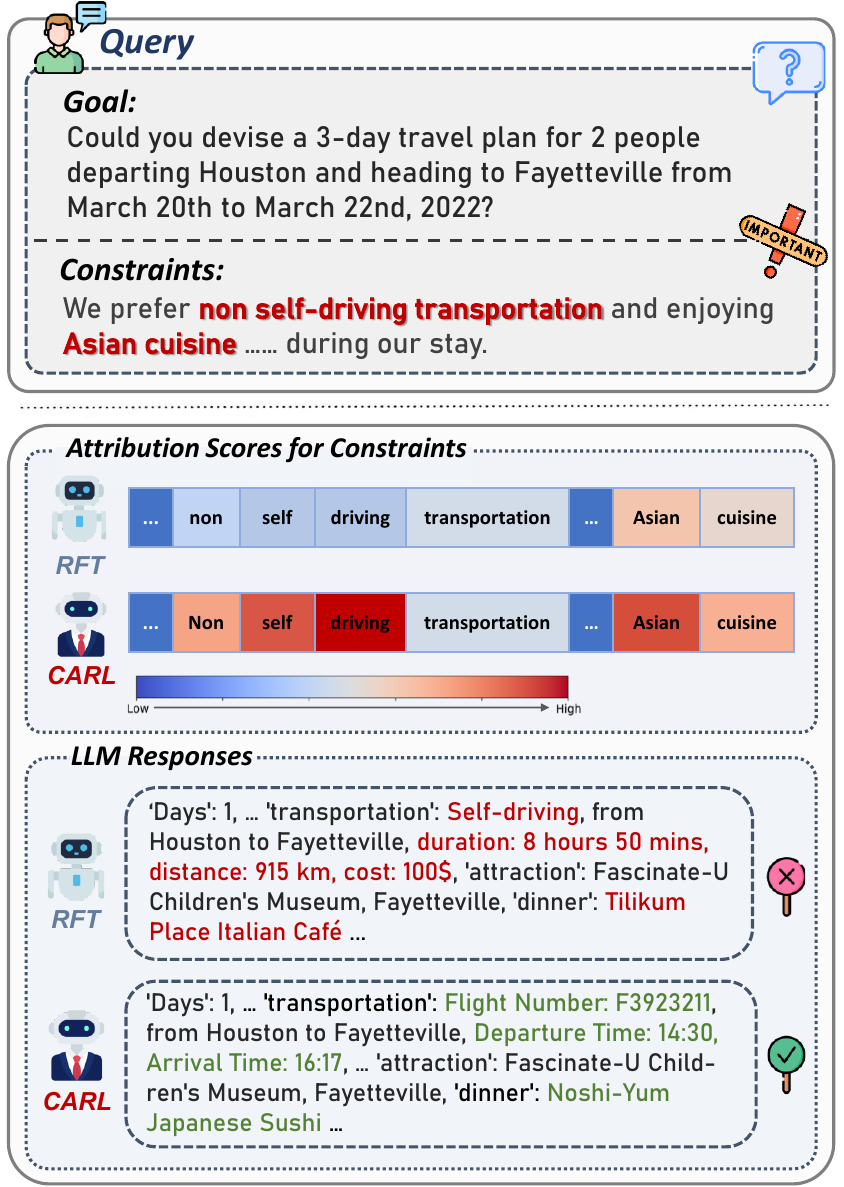}
  \caption{A typical case of planning tasks. The upper box demonstrates that a query can be decomposed into a goal and constraints, while the lower box shows that our CARL exhibits a higher focus on constraints, ultimately outperforming RFT in planning.}
  \label{fig:fig1}
\end{figure}

Despite these capabilities, LLMs consistently struggle to generate constraint-compliant plans in practice~\citep{wei2025plangenllms,huang2024understanding}. On the challenging real-world benchmark TravelPlanner~\citep{xie2024travelplanner}, DeepSeek-R1~\citep{guo2025deepseek}, a model renowned for its general reasoning performance, achieves a pass rate of only 12.2\%, falling significantly short of human-level performance. This disparity is not due to weak reasoning ability but rather reflects a fundamental limitation: \textbf{LLMs lack the capacity to systematically incorporate constraints into their generation process}. Empirical studies~\citep{xie2024revealing} corroborate this, revealing that LLMs frequently neglect constraints during planning and exhibit low attribution scores for constraint-related tokens in the input.

Existing approaches primarily circumvent this issue by offloading constraint reasoning to external scaffolds. Common paradigms include \textit{plan-then-execute}, which decomposes complex queries into a sequence of simpler subtasks~\citep{wang2023describe,singh2022progprompt}; \textit{step-by-step} frameworks that interleave planning with action execution in an iterative manner~\citep{wei2022chain}; and neuro-symbolic methods that translate natural language queries into formal planning representations (e.g., PDDL) for symbolic solvers~\citep{wu2022autoformalization,he2023solving}. While these methods yield performance gains, they rely heavily on closed-source models, external tools, or task-specific engineering. Crucially, they do not enhance the model's intrinsic understanding of constraints, thereby limiting generalizability and deployment in autonomous environments.

To address this, we propose \textbf{Constraint-Aware Reinforcement Learning (CARL)}, a general RL framework designed to explicitly strengthen LLMs' focus on constraints. CARL introduces a novel constraint-aware reward derived from the distributional shift of the model's outputs under constrained versus unconstrained inputs. By utilizing the KL divergence between log-probabilities in these two settings as a reward signal, CARL guides the model to integrate constraint signals more effectively. Unlike discrete task rewards, our continuous constraint-aware reward provides smoother optimization gradients and encourages meaningful exploration during planning failures.

CARL provides a novel and effective training framework for enhancing intrinsic planning competence. As illustrated in Figure~\ref{fig:fig1}, models trained with CARL exhibit significantly higher attribution scores for constraint tokens (e.g., non-self-driving transportation, Asian cuisine) compared to standard RFT, leading to improved plan validity. Furthermore, CARL is model-agnostic and compatible with a broad spectrum of RL algorithms, including on-policy methods like PPO~\citep{yu2022surprising} and GRPO~\citep{shao2024deepseekmath}, as well as off-policy approaches like DPO~\citep{rafailov2023direct}.

We evaluate CARL across three diverse planning benchmarks: BlocksWorld (block manipulation)~\citep{valmeekam2024llms}, TravelPlanner (travel planning)~\citep{xie2024travelplanner}, and T-Eval (tool use)~\citep{chen2023t}. The results demonstrate consistent and substantial improvements over RFT. Notably, CARL achieves a 56.1\% final pass rate on TravelPlanner—a +11.1\% gain over the baseline—outperforming state-of-the-art reasoning models such as o1-preview (10.0\%) and DeepSeek-R1 (12.2\%). Ablation studies and attribution analysis further confirm that these gains stem from a successfully acquired awareness of constraints. Our contributions are summarized as follows:

\begin{itemize} 
\item We propose CARL, a novel RL framework that systematically enhances LLMs' focus on constraints by modeling distributional shifts in output log-probabilities under constrained versus unconstrained inputs.

\item We design a learnable and interpretable constraint-aware reward mechanism that enables fine-grained control over constraint compliance. This formulation is highly generalizable, seamlessly extending to both on-policy and off-policy paradigms.

\item We demonstrate CARL’s effectiveness across diverse planning benchmarks, where it achieves substantial performance gains and significantly improved constraint focus, all without reliance on external solvers or top models. 
\end{itemize}

\section{Related Works}
\label{sec:related}
\subsection{Planning with LLMs}
\label{sec:related1}
Large Language Models (LLMs) have exhibited impressive potential in complex planning tasks, driven by their reasoning capabilities~\citep{yao2023react,kojima2022large,raman2024cape} and proficiency in tool utilization~\citep{qin2023toolllm,schick2023toolformer}. Leveraging LLMs' zero-shot generalization, early studies explored direct planning approaches~\citep{huang2022language,ahn2022can}, though these were often limited to simple, grounded scenarios. To tackle more complex problems, Chain-of-Thought (CoT) prompting was adopted to induce structured reasoning~\citep{wei2022chain}. More recently, tool-augmented frameworks have emerged to bolster planning reliability: some translate problems into formal logic for external solvers~\citep{liu2023llm+,xie2023translating,gundawar2024robust}, others integrate code snippets to handle dynamic "what-if" scenarios~\citep{li2023large}, and several implement iterative refinement loops utilizing task-specific verifiers or human feedback~\citep{kambhampati2024llms,chen2024autotamp}.

A common limitation of these approaches is that they treat LLMs as static components—either as high-level dispatchers or translators—rather than fundamentally improving their planning proficiency. Consequently, they externalize complex reasoning and constraint handling, creating dependencies on external tools and top models. While fine-tuning offers a direct pathway to acquire intrinsic skills, its application to planning tasks remains under-explored, particularly regarding the systematic integration of constraints during generation.

In contrast, we propose a novel RL paradigm that directly enhances the model's constraint-aware planning capabilities at the policy level. Instead of relying on external scaffolding, our method instills planning competence directly into the model via the constraint-aware reward, enabling superior performance and autonomy without dependence on external solvers, top models, or extensive prompt engineering.

\begin{figure*}[t]
\centering
    \includegraphics[width=\linewidth]{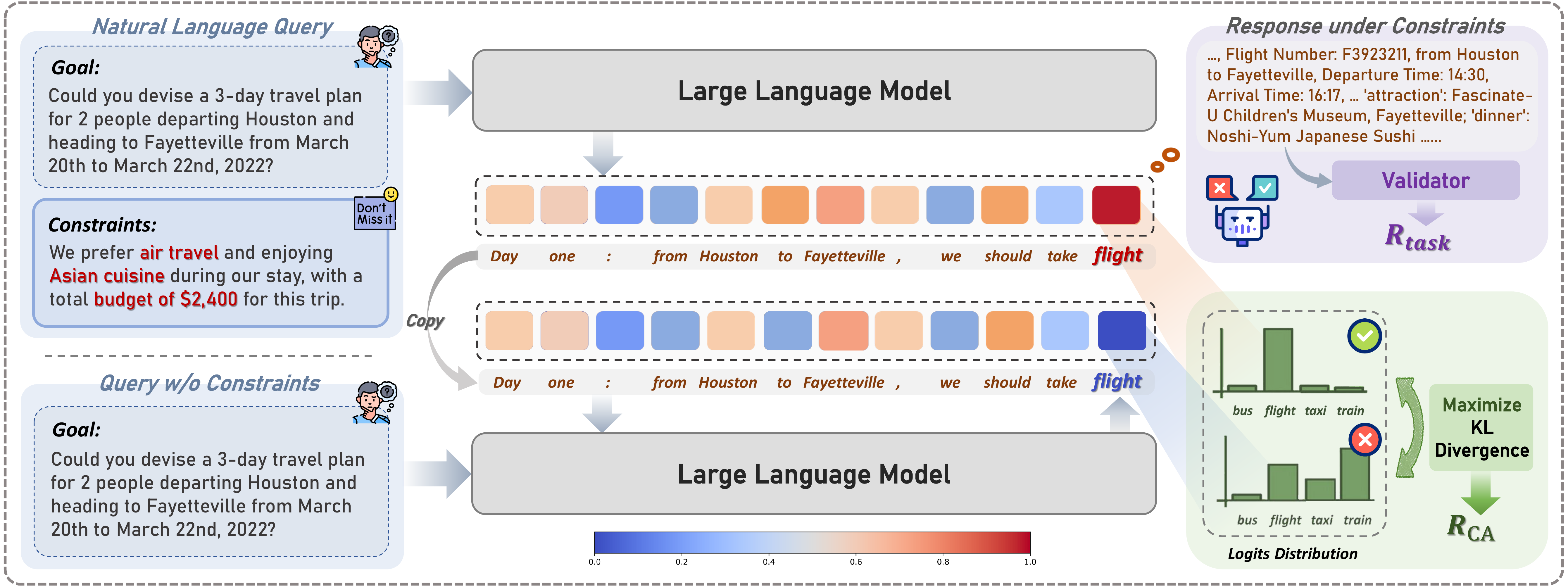}
    \caption{The framework of our proposed CARL. The reward is decomposed into two components: $R_{\text{task}}$ for achieving task-oriented objectives and $R_{\text{CA}}$ for sensitive adherence to constraints.}
    \label{fig:fig2}
\end{figure*}

\subsection{Reinforcement Learning for LLMs}
\label{sec:related2}
Reinforcement Learning from Human Feedback (RLHF)~\citep{kaufmann2024survey} adapts RL for LLMs, traditionally employing Proximal Policy Optimization (PPO)~\citep{yu2022surprising}. To improve stability and efficiency, methods like Direct Preference Optimization (DPO)~\citep{rafailov2023direct} and SimPO~\citep{meng2024simpo} have been developed, though often at the cost of on-policy performance. Recent advances such as Group Relative Policy Optimization (GRPO)~\citep{shao2024deepseekmath} and Reinforcement Learning with Online Optimization (RLOO)~\citep{ahmadian2024back} seek to balance performance with computational efficiency.

Parallel research has attempted to enhance input awareness through specialized rewards or preference pairs. For example, \citet{kiruluta2025self} utilize attention scores as reward signals to prioritize crucial tokens, while \citet{gu2024token,deng2024efficient} construct preference pairs from original and noise-perturbed images to strengthen visual anchoring. However, these methods are primarily limited to simple QA or visual tasks, and their efficacy in complex reasoning scenarios such as planning remains unverified.

In this work, we introduce a constraint-aware reward mechanism tailored for planning tasks. Our key innovation lies in quantifying constraint sensitivity through the distributional shift in log-probabilities between constrained and unconstrained conditions. This formulation is natively compatible with on-policy methods (PPO, GRPO) and can be seamlessly adapted to off-policy frameworks like DPO by transforming constraint signals into preference pairs, ensuring robust constraint compliance across training paradigms.

\section{Preliminaries}
\label{sec:preliminaries}
\subsection{Planning Task Formulation and Decomposition}
\label{sec:PTFD}
To faithfully address the challenges in planning tasks, as shown in Figure~\ref{fig:fig2}, a planning task query can be decomposed into two components: goal (the final target to achieve) and constraints (the conditions that need to be adhered to). On this basis, we define an unconstrained planning problem to more systematically examine the role of constraints in achieving reliable planning outcomes. Our method is designed to be broadly applicable across generic planning scenarios. Specifically, a planning task $\mathcal{P}$ is represented as an input sequence:
\begin{equation}
    x = (x_1, x_2, \dots, x_T) \in \mathbb{R}^{T \times d}
\end{equation}
where $T$ is the input length, and $d$ is the dimensionality of the embedding space. 

Constraints are an inherent part of all planning tasks and play a pivotal role in ensuring outcome correctness. In our framework, such constraints are identified and modeled through the following procedures:
\begin{itemize}
\item For tasks with explicitly stated constraints, such as \texttt{TravelPlanner}, we utilize the predefined constraints explicitly described within the task query.
\item For tasks with implicitly defined constraints, such as \texttt{T-Eval}, we employ constraint-extraction heuristics. Specifically, a lightweight prompt-based approach extracts and isolates the constraint-relevant portions of the input query, with full implementation details and sensitivity analysis provided in Appendix~\ref{sec:appendix_PCE}.

\end{itemize}

Formally, we denote the indices corresponding to the constraint tokens as $\mathcal{C} \subseteq \{1, 2, \dots, T\}$. The extracted sequence of constraint-specific tokens is then $x_{\mathcal{C}} = (x_t)_{t \in \mathcal{C}}$. The remaining subsequence, which represents the unconstrained portion of the planning task, is defined as:
\begin{equation}
    x_{\setminus \mathcal{C}} = \left( x_t \right)_{t \in \{1,\dots,T\} \setminus \mathcal{C}}
\end{equation}
While the unconstrained planning task $\mathcal{P}_{\setminus \mathcal{C}}$ may capture general goal-related elements of the task, it omits the crucial information encoded within constraints, potentially leading to incomplete or invalid solutions. This observation motivates our decomposition approach: constraints often act as the governing principles that disambiguate solutions and guarantee their feasibility. This distinction sets the stage for our proposed constraint-aware reinforcement learning framework.

\subsection{Group Relative Policy Optimization (GRPO)}
\label{sec:GRPO}

GRPO~\citep{shao2024deepseekmath} is an on-policy reinforcement learning algorithm. In the context of planning, consider a dataset $D$ containing datapoints consisting of inputs $x$. The GRPO learning objective with respect to the policy $\pi_{\theta}$ can be written as follows, where $\theta$ represents the parameters in a large language model:
\begin{equation}
\small
    \begin{split}
        &\mathcal{J}_{\text{GRPO}}(\theta) = \mathbb{E}_{[\{y_i\}_{i=1}^G \sim \pi_{\theta_{old}}(Y|x)]} 
            \frac{1}{G}\sum_{i=1}^G\frac{1}{|y_i|} \sum_{t=1}^{|y_i|} \Big\{ \\
        &\quad\quad \min \left[ r_{i,t}(\theta) \hat{A}_{i,t},\, 
            \text{clip} \left( r_{i,t}(\theta), 1 - \epsilon_{l}, 1 + \epsilon_{h} \right) \hat{A}_{i,t} \right] \\
        &\quad\quad - \beta \mathbb{D}_{KL}\left[\pi_{\theta} \| \pi_{ref}\right] \Big\} \\
        &\text{with} \; r_{i,t}(\theta) = 
            \frac{\pi_\theta(y_{i,t} | x, y_{i,<t})}{\pi_{\theta_{old}}(y_{i,t} | x, y_{i,<t})}
    \end{split}
    \label{eq:grpo}
\end{equation}
$G$ denotes the size of the group which contains multiple responses $Y$ sampled from the rollout policy $\pi_{\theta_{old}}$, corresponding to one input instance $x$. 
$\epsilon_l, \epsilon_h \in R$ are hyperparameters for clipping too large updates. The token-level advantage $\hat{A}_{i,t}$ is defined as the sequence-level reward normalized across the group.

\section{Methodology}
\label{sec:method}

\subsection{Overview}
\label{sec:overview}
To address the dual objectives of goal attainment and constraint satisfaction in planning, we reformulate the standard reinforcement learning (RL) objective by decomposing the overall reward signal $R$ into two components, as shown in Figure~\ref{fig:fig2}: a task-specific reward $R_{\text{task}}$ for achieving task-oriented objectives, and a constraint-aware reward $R_{\text{CA}}$ for sensitive adherence to constraints:
\begin{equation}
    R(x, y) = R_{\text{task}}(x, y) + \alpha R_{\text{CA}}(x, y)
    \label{eq:reward}
\end{equation}
where $\alpha \geq 0$ is a hyperparameter that regulates the relative importance of constraint adherence in the learning process. The trade-off between these components balances task achievement with constraint compliance.

At its core, our methodology embeds constraint-awareness into both the reward shaping and the policy optimization steps, detailed below.

\subsection{Constraint-Aware Reward Shaping}
\label{sec:CARS}
We introduce a novel reward shaping mechanism that explicitly integrates constraint-awareness into the learning objective. The task reward $R_{\text{task}}$ is computed as:
\begin{equation}
    R_{\text{task}}(x, y) = \mathbb{I}[y \in \mathcal{Y}_{\text{valid}}(x)]
\end{equation}
where $\mathbb{I}[\cdot]$ is an indicator evaluating to 1 if the generated output $y$ satisfies the task-specific success criterion (task achievement), and $\mathcal{Y}_{\text{valid}}(x)$ is the task-valid output space for input $x$. The constraint-aware reward $R_{\text{CA}}$ is defined as:
\begin{equation}
    R_{\text{CA}}(x, y) = \mathbb{D}_{\mathrm{KL}}[\pi_{\theta}(y | x) \;\|\; \pi_{\theta}(y | x_{\setminus\mathcal{C}})]
    \label{eq:CAR}
\end{equation}
Intuitively, the KL divergence captures the extent to which constraints influence model behavior, encouraging generation patterns that adhere to the constraint information. Combining this reward with the RL objective (e.g., GRPO) yields the complete CARL objective:
\begin{equation}
\small
    \begin{split}
        &\mathcal{J}_{\text{CARL}}(\theta) = \mathbb{E}_{[\{y_i\}_{i=1}^G \sim \pi_{\theta_{old}}(Y|x)]} 
            \frac{1}{G}\sum_{i=1}^G\frac{1}{|y_i|} \sum_{t=1}^{|y_i|} \Big\{ \\
        &\quad\quad \min \left[ r_{i,t}(\theta) \hat{A}_{i,t},\, 
            \text{clip} \left( r_{i,t}(\theta), 1 - \epsilon_{l}, 1 + \epsilon_{h} \right) \hat{A}_{i,t} \right] \\
        &\quad\quad - \beta \mathbb{D}_{KL}\left[\pi_{\theta} \| \pi_{ref}\right] \\
        &\quad\quad \textcolor{MyDarkOrange}{+\, \alpha \mathbb{D}_{\mathrm{KL}}[\pi_{\theta}(y_i | x) \;\|\; \pi_{\theta}(y_i | x_{\setminus\mathcal{C}})]} \Big\}
    \end{split}
    \label{eq:objective}
\end{equation}
where $i$ indexes the $i$-th rollout response. $\alpha$ and $\beta$ are weighting coefficients used for constraint-aware reward and KL penalty ($\mathbb{D}_{KL}\left[\pi_{\theta} || \pi_{ref}\right]$). We then compare the GRPO-version training dynamics and efficiency of CARL and RFT, as shown in Appendix~\ref{sec:appendix_C_RFT}.

\subsection{Constraint-Aware Direct Preference Optimization}
\label{sec:CADPO}
To adapt our constraint-aware paradigm to preference-based learning, we propose a novel extension of Direct Preference Optimization (DPO) that injects constraint focus through strategic preference construction. Our key innovation lies in generating contrastive responses under constraint ablation to create informative preference pairs.

Given a standard response $y \sim \pi(\cdot|x)$ and its constraint-ablated counterpart $y_{\setminus\mathcal{C}} \sim \pi(\cdot|x_{\setminus\mathcal{C}})$, the preference dataset is defined as:
\begin{equation}
    \mathcal{D}_{\text{CA}} = \left\{(x^{(i)}, y^{(i)}, y_{\setminus\mathcal{C}}^{(i)}) \,|\, y^{(i)} \in \mathcal{Y}_{\text{valid}}(x^{(i)})\right\}_{i=1}^N
\end{equation}
The preference-based DPO objective is then augmented to reflect the impact of constraint ablation:
\begin{equation}
    \mathcal{L}_{\text{CA-DPO}} = -\mathbb{E}_{\mathcal{D}_{\text{CA}}} \left[ \log \sigma\left( \beta \Delta(x,y,y_{\setminus\mathcal{C}}) \right) \right]
\end{equation}
The constrained advantage function $\Delta$ is computed as:
\begin{equation}
    \Delta(x,y,y_{\setminus\mathcal{C}}) = \log\frac{\pi_\theta(y|x)}{\pi_{\text{ref}}(y|x)} - \log\frac{\pi_\theta(y_{\setminus\mathcal{C}}|x)}{\pi_{\text{ref}}(y_{\setminus\mathcal{C}}|x)}
\end{equation}
This formulation introduces an implicit reward margin that quantifies constraint influence on policy outputs, effectively incentivizing the model to generate constraint-compliant responses through optimization. The corresponding results are shown in Figure~\ref{fig:fig4}.

\begin{table*}[t]

\centering

\resizebox{\textwidth}{!}{
\begin{tabular}{l c cc cc c ccc} 
\toprule
\multirow{3}{*}{\textbf{Model}} & \multirow{3}{*}{\textbf{BlocksWorld}} & \multicolumn{5}{c}{\textbf{TravelPlanner}} & \multicolumn{3}{c}{\textbf{T-Eval}}  \\
\cmidrule(lr){3-7} \cmidrule(lr){8-10}
& & \multicolumn{2}{c}{\textbf{Commonsense}} & \multicolumn{2}{c}{\textbf{Hard}} & \multirow{2}{*}{\textbf{Final}} & \multirow{2}{*}{\textbf{Precision}} & \multirow{2}{*}{\textbf{Recall}} & \multirow{2}{*}{\textbf{F1-score}} \\
\cmidrule(lr){3-4} \cmidrule(lr){5-6}

& & \textbf{Micro} & \textbf{Macro} & \textbf{Micro} & \textbf{Macro} & & & & \\
\midrule
GPT-4o      & 42.4 & 84.7 & 31.1 & 53.6 & 31.1 & 7.8  & 90.4 & 86.4 &  87.5 \\
o1-preview  & 97.8 & 79.6 & 15.0 & 41.9 & 37.8 & 10.0 & 90.0 & 86.5 & 87.4 \\
DeepSeek-V3       & 44.8 & 80.3 & 17.2 & 30.5 & 13.9 & 2.2  & 91.1 & 87.4 & 88.5 \\
DeepSeek-R1       & 98.2 & 80.6 & 22.2 & 51.7 & 41.7 & 12.2  & 90.2 & 87.2 & 87.8 \\
Qwen2.5-72B-Instruct & 13.8 & 82.3 & 16.7 & 32.6 & 22.8 & 6.1 & 92.2 & 88.1 & 89.2 \\
QwQ-32B & 88.8 & 74.9 & 6.1 & 41.4 & 32.8 & 4.4 & 88.3 & 84.6 & 85.6 \\
Llama-3.1-70B-Instruct & 21.6 & 82.8 & 18.9 & 33.1 & 16.1 & 2.2 & 85.4 & 81.9 & 83.0 \\

Llama-3.1-8B-Instruct & 0.6 & 60.1 & 0.0 & 7.9 & 2.8 & 0.0 & 81.5 & 76.5 & 78.9 \\

DeepSeek-R1-Distill-Llama-8B & 1.4 & 61.2 & 0.0 & 0.0 & 0.0 & 0.0 & 81.8 & 79.3 & 79.4 \\
Qwen3-8B    & 31.2 & 72.7 & 7.8  & 34.8 & 27.8 & 2.2  & 86.6 & 83.5 & 84.2  \\
\midrule

DeepSeek-R1-Distill-Llama-8B (RFT)        & 42.0 & 80.8 & 25.0 & 36.2 & 19.4 & 5.6 & 88.0 & 86.0 & 86.4 \\
DeepSeek-R1-Distill-Llama-8B (Ours)       & 52.6 & 81.1 & 32.1 & 42.9 & 28.9 & 11.7 & 88.4 & \textbf{88.6} & \underline{87.5} \\
Qwen3-8B (RFT)        & \underline{73.8} & \underline{96.3} & \underline{74.4} & \underline{65.7} & \underline{48.9} & \underline{45.0} & \underline{88.6} & 88.2 &  87.3 \\
Qwen3-8B (Ours)       & \textbf{77.2} & \textbf{97.3} & \textbf{81.7} & \textbf{73.1} & \textbf{59.4} & \textbf{56.1} & \textbf{89.5} & \underline{88.5} & \textbf{88.1} \\
\bottomrule
\end{tabular}
}
\caption{Results on planning benchmarks. Unless otherwise specified, both RFT and CARL are implemented based on GRPO. The best and second-best results are \textbf{bold} and \underline{underlined}.}
\label{tab:1}
\end{table*}

\section{Experiments}
\label{sec:experiments}
\subsection{Datasets and Settings}
\label{sec:datasets}
\paragraph{Datasets.}
We evaluate CARL across three complementary planning benchmarks that collectively cover \textit{classical symbolic planning}, \textit{real-world constrained decision-making}, and \textit{tool-mediated planning}:
\begin{itemize}
    \item \textbf{BlocksWorld}~\citep{valmeekam2024llms} is a formal symbolic planning environment with well-defined action schemas and \textbf{static} constraints. Given an initial block configuration and a goal state, models must generate action sequences that strictly adhere to the physical constraints specified in the prompt. This benchmark provides a controlled setting to isolate and evaluate core constraint-handling capabilities.

    \item \textbf{TravelPlanner}~\citep{xie2024travelplanner} presents a real-world travel planning challenge where models must generate plans based on provided information and user queries, aligning with commonsense and the hard constraints specified in the queries. Unlike the static nature of BlocksWorld, the hard constraints in TravelPlanner are \textbf{dynamic}, as they need to be inferred from the query and satisfied through item selection.

    \item \textbf{T-Eval}~\citep{chen2023t} is a fine-grained benchmark assessing LLMs' tool-use ability across multiple evaluation aspects. In this work, we primarily focus on its planning task. As noted in Sec.~\ref{sec:PTFD}, T-Eval is characterized by \textbf{implicit} constraints, which are embedded within the input queries. We then use GPT-4o with a lightweight prompt to extract these constraints.
    
\end{itemize}

We follow the official partitions for BlocksWorld and TravelPlanner, using 100 and 45 samples for training, and 500 and 180 samples for testing, respectively. For T-Eval, we randomly select 128 samples from the 553-sample evaluation set for training, with the remaining samples for testing.

\paragraph{Metrics.}
Accuracy is used for evaluating BlocksWorld, whereas precision, recall, and F1-score are used for T-Eval\footnote{In T-Eval, the reported precision, recall, and F1-score are the arithmetic means of the per-sample scores.}. For TravelPlanner, we employ a multi-faceted evaluation framework that separates commonsense from hard constraints, with two complementary metrics reported for each:
\begin{itemize}
    \item \textit{Micro pass rate}: The ratio of successfully satisfied constraints to total constraints of that type.
    \item \textit{Macro pass rate}: The ratio of plans satisfying all constraints of that type to total plans.
\end{itemize}

Finally, we use the \textit{final pass rate} as the proportion of plans satisfying all constraints, which corresponds to the \textit{macro pass rate} when considering all constraints collectively, representing the ultimate planning success metric.

\paragraph{Implementation Details.}
Our reinforcement learning framework is implemented based on Verl~\citep{sheng2025hybridflow}.
Unless otherwise specified, we adopt Qwen3-8B as the base model to balance performance and training efficiency, and use GRPO for optimization. Training is conducted on a single node with 8 A100 GPUs, and each step samples a batch of 64 queries with 8 rollouts. The weighting coefficients $\alpha$ and $\beta$ (see Equation~\ref{eq:objective}) are both set to 0.001. Additional training details are provided in Appendix~\ref{sec:appendix_ID_GRPO}.

\subsection{CARL Performance}
\label{sec:MR}
\paragraph{Results on Planning Benchmarks.}
Table~\ref{tab:1} summarizes the main evaluation results across all planning benchmarks, highlighting the comparative performance of CARL against state-of-the-art LLMs and RFT. There are three key takeaways:

First, {CARL significantly boosts the planning capabilities of Qwen3-8B and DeepSeek-R1-Distill-Llama-8B.} For example, with fewer than 128 training queries, CARL significantly boosts Qwen3-8B’s planning performance, achieving an absolute improvement of 53.9\% on TravelPlanner (from 2.2\% to 56.1\%). It consistently outperforms RFT and significantly surpasses state-of-the-art reasoning-oriented models such as o1-preview (10.0\%) and DeepSeek-R1 (12.2\%). Furthermore, CARL enables the performance of the 14B model comparable to top models on BlocksWorld, as shown in Figure~\ref{fig:fig3} and Table~\ref{tab:2}.

Second, {CARL effectively mitigates constraint neglects or violations in multi-constraint planning tasks.} While baseline models often perform well in terms of \textit{micro pass rate} (i.e., the proportion of individual constraints satisfied), their \textit{macro pass rate} (i.e., the proportion of plans satisfying all constraints) remains low. This indicates that performance bottlenecks arise from constraint neglect or violation in multi-constraint scenarios, rather than inherent difficulty in satisfying individual constraints. By enhancing the model’s focus on constraints, CARL substantially mitigates such issues, leading to improved \textit{macro} and \textit{final pass rates}.

Last, {CARL outperforms RFT in handling dynamic hard constraints.} Although both CARL and RFT support the learning of static commonsense constraints, CARL proves more effective at handling dynamic constraints—hard constraints that vary across instances—resulting in higher overall task success rates than RFT.

To summarize, CARL consistently demonstrates superior planning performance and robustness across benchmarks and models. Additional experimental results, including comparisons with SFT, prompting, and agent-based methods, as well as generalization performance on other benchmarks, are presented in Appendix~\ref{sec:appendix_MRPB}.

\begin{figure}[t]
\centering
  \includegraphics[width=\columnwidth]{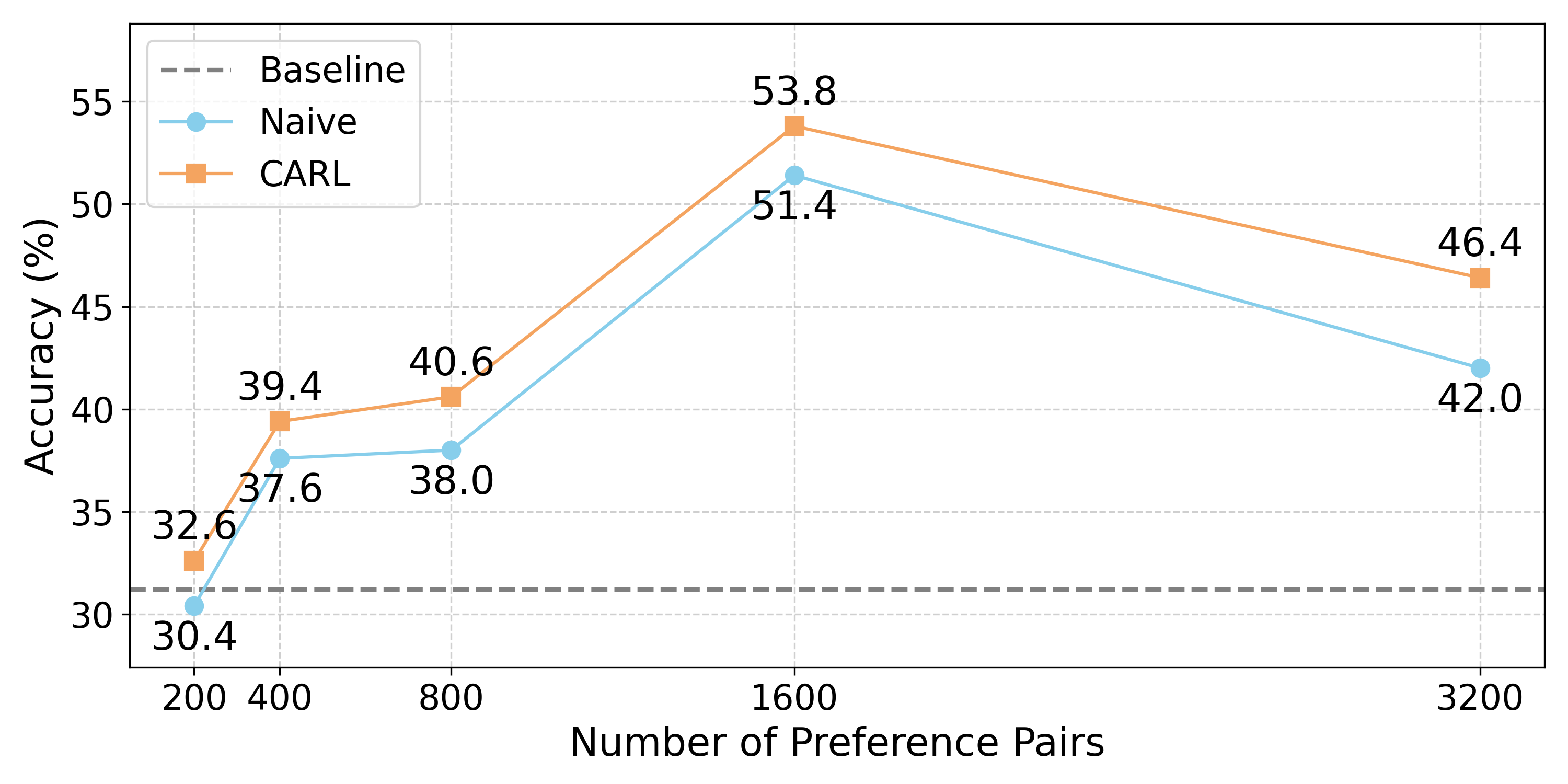}
  \caption{Performance comparison of DPO and CA-DPO with different preference pairs on BlocksWorld.}
  \label{fig:fig4}
\end{figure}

\paragraph{Adaptation to Other RL Methods.}
Our proposed CARL framework can be seamlessly extended to other representative on-policy and off-policy reinforcement learning methods, such as PPO and DPO. Table~\ref{tab:3} summarizes the performance of these CARL implementations, consistently showing superior results compared to their naive counterparts.

The implementation of CARL on PPO is similar to that on GRPO; however, for DPO, we begin by constructing preference pairs using responses generated from the base model. Specifically, we input constrained queries and construct preference pairs based on the correctness of the responses for naive DPO. For CA-DPO, we input both constrained and unconstrained queries. The correct responses generated for constrained queries are selected as positive samples, while the responses generated for unconstrained queries are randomly chosen as negative samples, which are then utilized to construct preference pairs. For both methods, we maintain the same total number of preference pairs and use identical parameter settings for optimization. Experimental results, shown in Figure~\ref{fig:fig4}, reveal that CA-DPO consistently outperforms naive DPO. With as few as 200 preference pairs, CARL enables the model to be aware of constraints and slightly improves planning performance. As the number of preference pairs increases, the diversity of data expands, leading to a continuous improvement in the model’s planning capabilities. However, when the number of preference pairs reaches 3,200, the model's performance begins to decline. We attribute this to a bottleneck in data diversity among the preference pairs generated by the base model. We think that introducing additional data sources in future work could further enhance the effectiveness of DPO. For more details about DPO, please refer to Appendix~\ref{sec:appendix_ID_DPO}.

\begin{figure}[t]
\centering
  \includegraphics[width=\columnwidth]{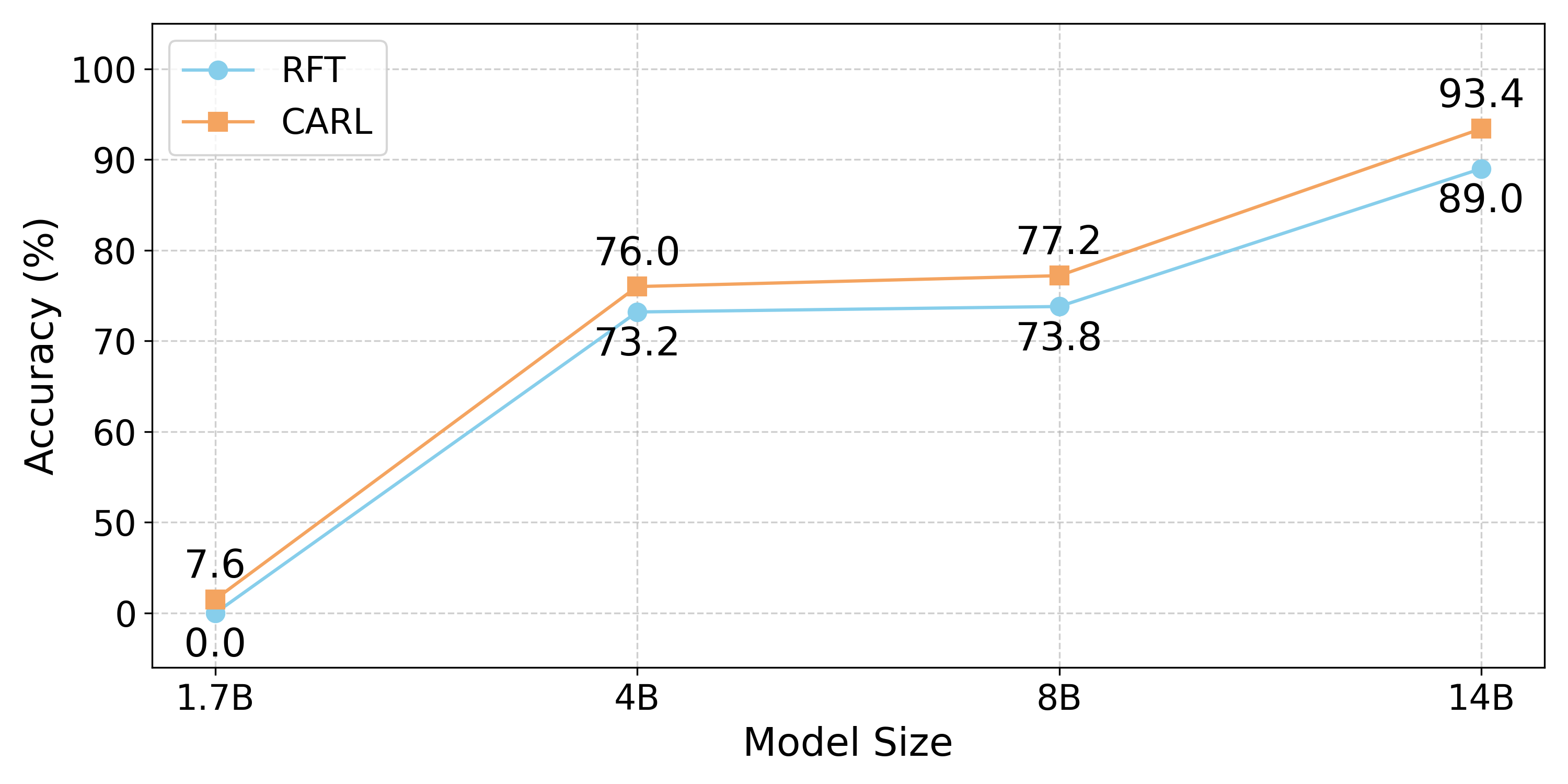}
  \caption{Performance comparison of RFT and CARL with different size Qwen3 models on BlocksWorld.}
  \label{fig:fig3}
\end{figure}

\begin{table}[t]
  \centering
  
  \resizebox{\columnwidth}{!}{
  \begin{tabular}{lcccc}
\toprule
\multirow{2}{*}{Model} & \multicolumn{2}{c}{PPO} & \multicolumn{2}{c}{DPO} \\
\cmidrule(lr){2-3} \cmidrule(lr){4-5}
                       & Naive        & CARL        & Naive       & CARL       \\
\midrule
Qwen3-8B               & 78.4         & 81.6        & 51.4        & 53.8       \\
\bottomrule
\end{tabular}
}
  \caption{Performance comparison of RFT and CARL with different RL methods on BlocksWorld.}
  \label{tab:3}
\end{table}

\subsection{Ablation Studies}
\label{sec:Ablation_Studies}
We demonstrate the effectiveness of our constraint-aware reward through ablation studies of model size, reward masking, and reward calculation strategies. Additionally, we investigate the sensitivity of the constraint-aware reward weight (i.e., hyperparameter $\alpha$) in Appendix~\ref{sec:appendix_sacarw}.

\paragraph{Model Size.}
To investigate the impact of model scale, we apply CARL to four Qwen3 models of increasing size. Results in Figure~\ref{fig:fig3} highlight two key effects:

\begin{itemize}
\item \textbf{Exploration Guidance:} For small models lacking initial planning ability (e.g., Qwen3-1.7B), RFT fails due to the absence of task rewards during rollout. In contrast, CARL’s constraint-aware reward provides smoother gradients, enabling effective exploration and gradual acquisition of planning skills through constraint focus.

\item \textbf{Constraint Grounding:} For models with basic planning ability, CARL’s advantage over RFT grows with scale, as larger models better leverage constraint signals to activate stronger reasoning patterns and boost planning performance.

\end{itemize}

\begin{table}[t]
  \centering
  
  \resizebox{\columnwidth}{!}{
  \begin{tabular}{lcccc}
    \toprule
    Model & None & Both & Goal & Constraint \\
    \midrule
    Qwen3-8B               & 73.8         & 71.2        & 70.4         & 77.2           \\
    \bottomrule
    \end{tabular}
    }
  \caption{Ablation study of reward masking strategy on BlocksWorld. ``Both'' denotes masking both goals and constraints.}
  \label{tab:4}
\end{table}

\paragraph{Reward Masking Strategy.} 
We assess the reward design by extending masking targets to goals or both components. As shown in Table~\ref{tab:4}, these variants perform worse than RFT (None). We hypothesize that removing goals disrupts the core task definition, leading to incoherent outputs and noisy supervision. In contrast, queries without constraints can still produce valid plans, allowing the reward to suppress constraint neglect.

\paragraph{Reward Calculation Strategy.}
To investigate the impact of the reward calculation strategy on training performance and dynamics, we measure the discrepancy between the log-probabilities in Equation~\ref{eq:CAR} using four metrics: \textit{difference} (dif), \textit{absolute difference} (abs), \textit{mean squared error} (mse), and \textit{low-variance kl divergence} (low\_var\_kl, which we used in our other experiments). As shown in Table~\ref{tab:AS_RCS}, all strategies facilitate stable training, with the notable exception of mse, which leads to collapse. The impact of reward calculation strategy on training dynamics is shown in Appendix~\ref{sec:appendix_IRCSTD}.

\subsection{Attribution Analysis}
\label{sec:AA}

\begin{figure}[t]
\centering
  \includegraphics[width=\columnwidth]{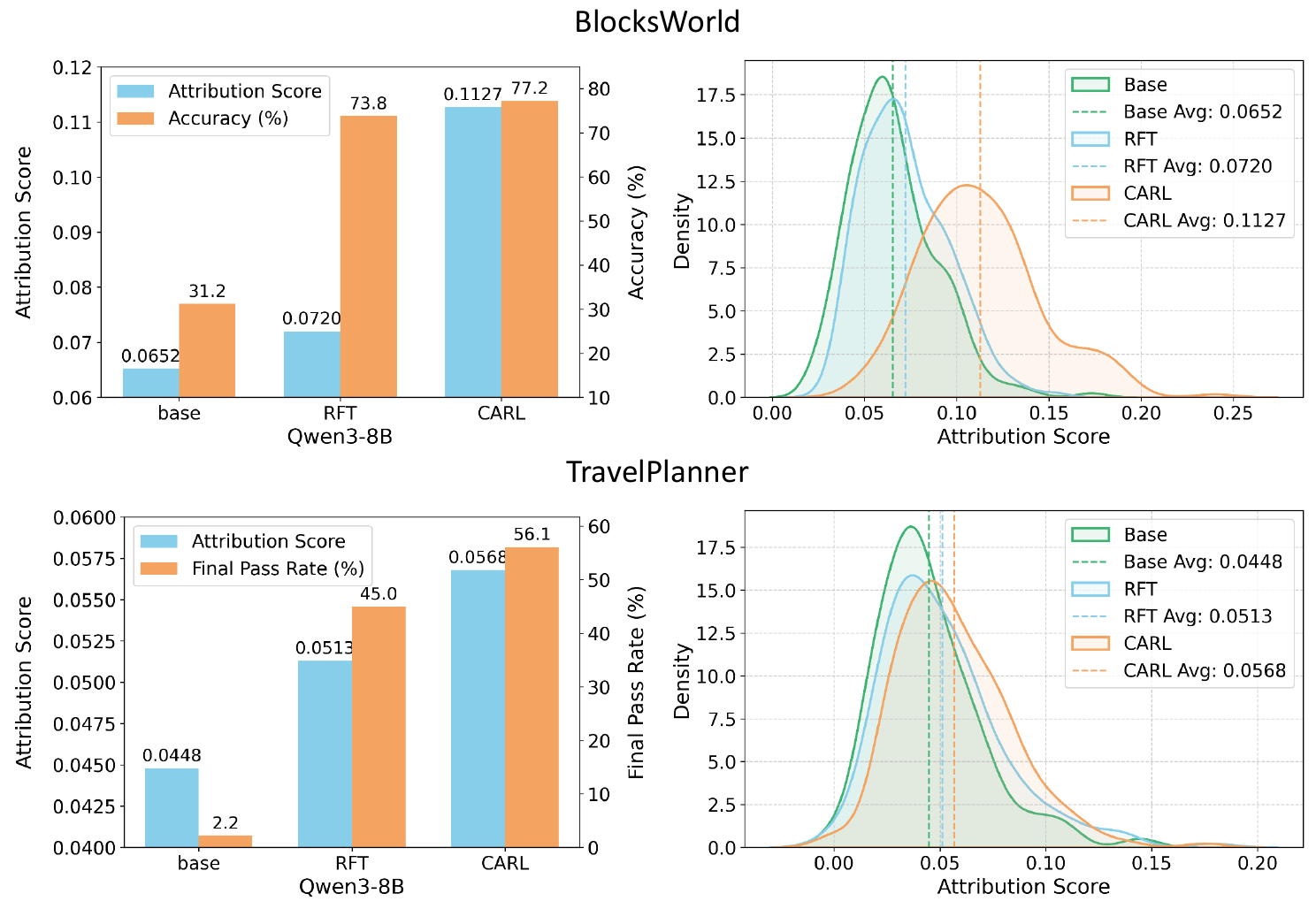}
  \caption{Attribution analysis on BlocksWorld and TravelPlanner. The left simultaneously presents the average attribution scores and performance of different models, while the right illustrates score distributions through Kernel Density Estimation.}
  \label{fig:fig5}
\end{figure}

To understand the source of planning performance improvements, we conduct an attribution analysis on BlocksWorld and TravelPlanner. We compare the base model with versions fine-tuned using RFT and CARL, focusing on the mean and distribution of attribution scores for constraint-related inputs (all scores are normalized by the response length). As shown in Figure~\ref{fig:fig5}, better task performance consistently correlates with higher attribution scores. Compared to RFT, CARL further enhances the model’s focus on constraints, leading to stronger performance.  
Importantly, this improvement is not driven by a few outliers but reflects a consistent shift in the overall distribution, indicating that CARL's gains stem from a general enhancement in constraint sensitivity. Additional details and the case study are provided in Appendix~\ref{sec:appendix_DAA}.

\begin{table}[t]
  \centering
  
  \resizebox{\columnwidth}{!}{
  \begin{tabular}{lcccc}
    \toprule
    Model & dif & abs & mse & low\_var\_kl \\
    \midrule
    Qwen3-8B        & 76.6       & 76.0                 & 0.2         & 77.2           \\
    \bottomrule
    \end{tabular}
    }
  \caption{Ablation study of reward calculation strategy on BlocksWorld. ``dif'', ``abs'', ``mse'', and ``low\_var\_kl'' denote different discrepancy metrics.}
  \label{tab:AS_RCS}
\end{table}

\vspace{-0.17em} 

\section{Conclusion}
\label{sec:conclusion}
In this paper, we introduced CARL, a novel RL framework designed to fundamentally enhance the intrinsic constraint sensitivity of LLMs in planning tasks. By leveraging the distributional divergence between model outputs under constrained and unconstrained inputs, CARL formulates an effective reward signal to improve constraint compliance without reliance on external solvers or task-specific engineering. Extensive experiments across diverse benchmarks demonstrate that CARL significantly outperforms both standard RFT baselines and state-of-the-art reasoning models. Furthermore, our analysis confirms that CARL induces a stronger focus on constraints while exhibiting robust generalization across different RL algorithms and model scales. Ultimately, this work establishes a scalable pathway for developing autonomous language agents capable of reliable, constraint-aware planning.

\section*{Limitations}
First, regarding training efficiency, CARL incurs a marginal computational overhead due to the additional forward pass for unconstrained queries. However, as detailed in Appendix~\ref{sec:appendix_TE}, this cost is significantly lower than that of scaling up rollouts to achieve comparable gains. Second, our evaluation is currently confined to single-agent benchmarks. Extending CARL to complex combinatorial planning scenarios, such as multi-robot collaboration, remains a direction for future work.

\section*{Acknowledgements}
This work was supported by the National Natural Science Foundation of China under Grant 42394060 and 42394064, Ant Group Research Fund, and Information Technology Center and State Key Lab of CAD\&CG, Zhejiang University.

\bibliography{acl2026_conference}
\clearpage
\appendix
\section{Prompt used for Constraint-Extraction}
\label{sec:appendix_PCE}
We herein present the prompt designed to extract implicit constraints in T-Eval, as referenced in Sec.~\ref{sec:PTFD}, along with its effect on T-Eval and a sensitivity analysis of the extraction quality.

\subsection{Full Prompt}
\label{sec:appendix_FP}
We use a simple prompt to extract and isolate the constraint-relevant portions of the input query for T-Eval, as shown below:
\\

\lstset{
    numbers=none, 
    backgroundcolor=\color[RGB]{245,245,245},
    breaklines=true,
    breakindent=0pt,
    basicstyle=\ttfamily\small,
    frame=trbl,
    frameround = tttt,
}\begin{lstlisting}
You are an expert at simplifying user queries by removing specific constraints while preserving the core intent.

Your task is to remove detailed constraints and specific requirements from the user query, keeping only the main objectives and essential context.

Rules:
1. Keep the main purpose and core actions
2. Remove specific numbers, quantities, limits
3. Remove specific dates, times, or temporal constraints
4. Remove detailed specifications or precise requirements
5. Maintain the overall structure and flow of the original query
6. Keep professional context and role descriptions
7. The output should be a simplified version that captures the essence without the fine-grained constraints

Now process the following:
Original user query:
{user_prompt}

Simplified query (remove constraints but keep core intent):
\end{lstlisting}

\subsection{Effect on T-Eval}
We use two cases to demonstrate the effect of our prompt on T-Eval, with the extracted implicit constraints underlined, as shown below:
\\
\lstset{
    numbers=none, 
    backgroundcolor=\color[RGB]{245,245,245},
    breaklines=true,
    breakindent=0pt,
    basicstyle=\ttfamily\small,
    frame=trbl,
    frameround = tttt,
    escapeinside={(*@}{@*)} 
}\begin{lstlisting}
(*@\textbf{[Case 1]:}@*)

(*@\textbf{Constrained Queries:}@*)
As the office manager, I need to find a meeting room (*@\underline{that is available for the}@*) (*@\underline{next two hours}@*) for a team meeting today. Once an available room is found, please book it (*@\underline{for the specified duration,}@*) (*@\underline{starting from the current time.}@*)

(*@\textbf{Unconstrained Queries:}@*)
As the office manager, I need to find a meeting room for a team meeting and book it.

(*@\textbf{[Case 2]:}@*)

(*@\textbf{Constrained Queries:}@*)
I am writing a research paper on quantum computing, and I need information about the (*@\underline{first}@*) author of the articles. Please find articles related to quantum computing and provide me with the meta information of the (*@\underline{first three}@*) articles. Lastly, I need to know if there are any meeting rooms available tomorrow (*@\underline{from 2:00 PM to 4:00 PM.}@*)

(*@\textbf{Unconstrained Queries:}@*)
I am writing a research paper on quantum computing, and I need information about the author of the articles. Please find articles related to quantum computing and provide me with the meta information of the articles. Lastly, I need to know if there are any meeting rooms available tomorrow.
\end{lstlisting}

\begin{figure}[t]
\centering
  \includegraphics[width=\columnwidth]{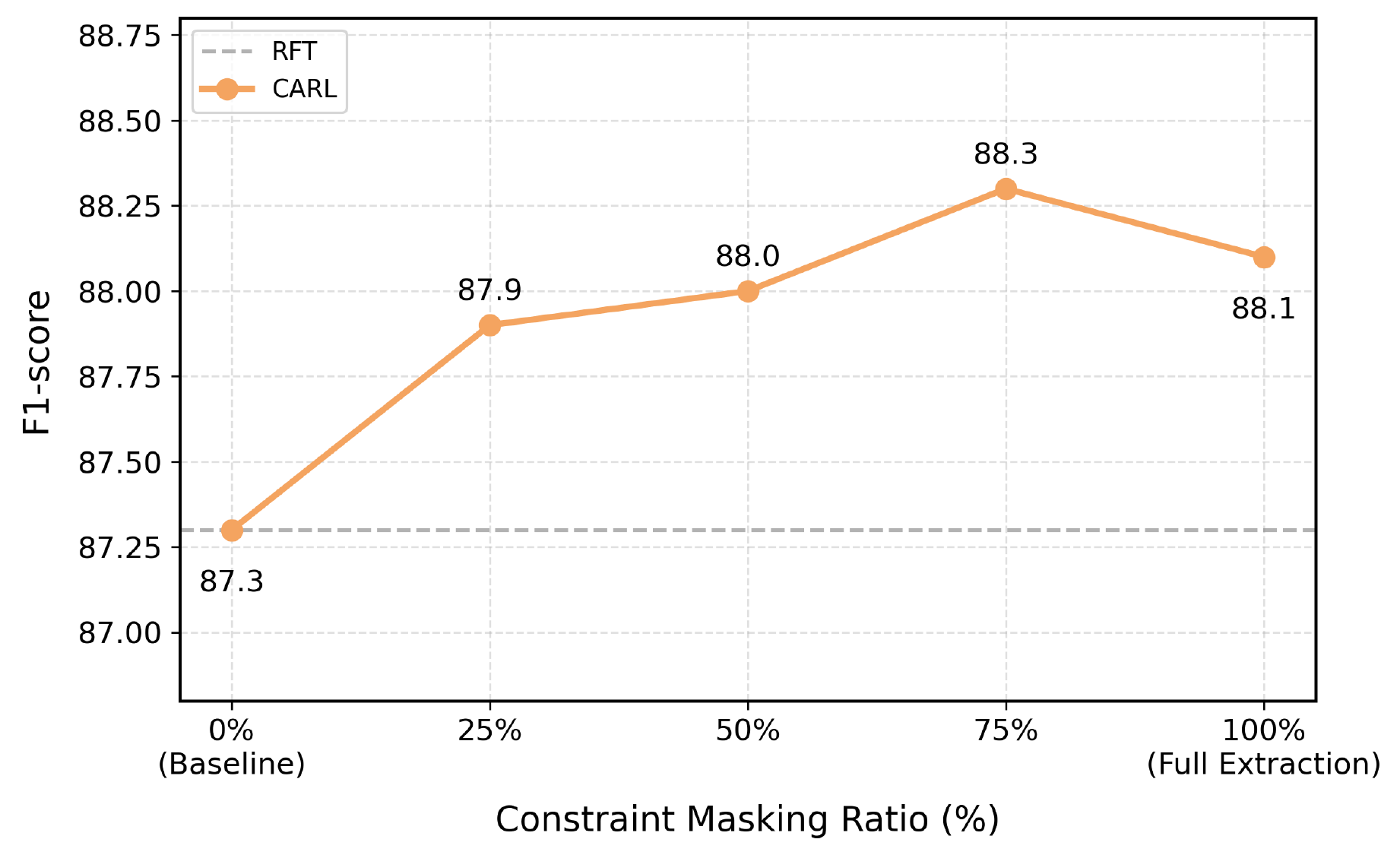}
  \caption{Sensitivity analysis of constraint extraction quality on T-Eval.}
  \label{fig:fig_saceq}
\end{figure}

\subsection{Sensitivity Analysis of Constraint Extraction Quality}

We herein investigate the sensitivity of the CARL framework to the quality of the constraint-extraction prompt. Specifically, we examine whether the performance gain relies on the exhaustive removal of constraint information from the unconstrained baseline, or if the model can acquire constraint awareness effectively from partial contrastive signals.

\begin{figure*}[t]
\centering
    \includegraphics[width=\linewidth]{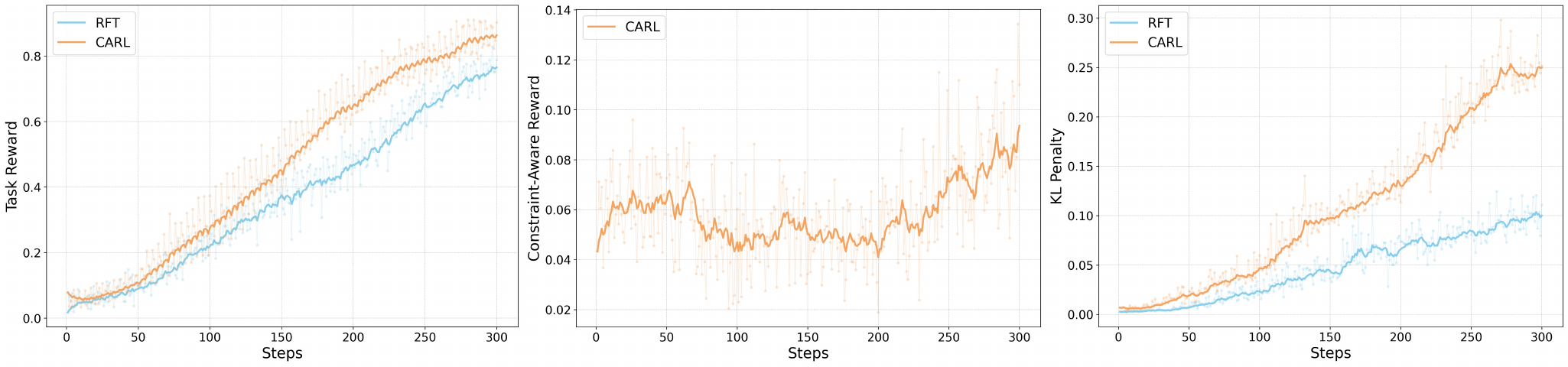}
    \caption{Comparison of the training dynamics on TravelPlanner. Solid lines indicate exponential moving averages of the data.}
    \label{fig:fig6}
\end{figure*}

\paragraph{Settings.}
We introduce the \textit{Constraint Masking Ratio} ($\rho$), defined as the proportion of constraint-related tokens identified and masked during the construction of the unconstrained query $x_{\setminus \mathcal{C}}$. Using the full constraint set extracted by the prompt in Appendix~\ref{sec:appendix_FP} as the pseudo-ground truth ($\rho=100\%$), we simulate varying levels of extraction quality by randomly downsampling the masked constraints at ratios of $\{0\%, 25\%, 50\%, 75\%, 100\%\}$. Note that $\rho=0\%$ corresponds to the standard RFT baseline, as it yields no contrastive signal. All experiments are conducted on the T-Eval benchmark using the Qwen3-8B model.

\paragraph{Results.}
The F1-scores on T-Eval across different masking ratios are illustrated in Figure~\ref{fig:fig_saceq}. Starting from the RFT baseline ($\rho=0\%$) with an F1-score of 87.3, applying a minimal masking ratio of $\rho=25\%$ results in a significant performance jump to 87.9. As $\rho$ increases, the performance exhibits a robust plateau with stochastic fluctuations: $\rho=50\%$ yields 88.0, while $\rho=75\%$ reaches an incidental peak of 88.3, before settling at 88.1 with full extraction ($\rho=100\%$).

\paragraph{Analysis.}
These results demonstrate that CARL is highly robust to the completeness of constraint extraction. A mere 25\% masking ratio captures the majority of the performance gain (0.6 out of 0.8), indicating that the model learns the \textit{general mechanism} of attending to constraints via partial contrastive signals, rather than relying on precise masking every constraint token. Furthermore, the fluctuation observed between 50\% and 100\% suggests that once sufficient contrast is established, further optimization of the extraction prompt yields marginal or noisy returns. This confirms that CARL enables robust constraint-aware planning without strict dependency on carefully designed extraction prompts.

\section{Comparison with RFT}
\label{sec:appendix_C_RFT}
We herein compare the GRPO-version training dynamics and efficiency of CARL and RFT.

\subsection{Training Dynamics}
\label{sec:appendix_TD}
Since we maximize a KL divergence that is theoretically unbounded, the model may ``hack'' the constraint-aware reward, eventually leading to performance collapse. The KL penalty, however, can mitigate the risk of reward hacking by constraining the magnitude of policy updates. As illustrated in Figure~\ref{fig:fig6} on TravelPlanner, although CARL exhibits a higher KL penalty compared to RFT—reflecting more significant policy updates driven by the additional constraint supervision—the training process remains stable. The task reward of CARL consistently surpasses that of RFT, showing faster convergence and higher final performance. Meanwhile, the constraint-aware reward fluctuates within a reasonable range without divergence. This demonstrates that the KL penalty effectively balances the optimization, allowing CARL to leverage constraint-aware signals for superior planning capabilities without succumbing to reward hacking.

\begin{table}[t]
  \centering
  \begin{tabular}{lccc}
\toprule
Method                  & Rollouts & Time per Step & Accuracy \\
\midrule
\multirow{2}{*}{RFT}    & 8         & 406.2      & 73.8     \\
                        & 16        & 674.2      & 75.0     \\
\midrule
CARL                    & 8         & 445.7      & 77.2     \\
\bottomrule
\end{tabular}
  \caption{Efficiency analysis with different rollouts on BlocksWorld. Results are reported in terms of training time per step (s) and accuracy (\%).}
  \label{tab:5}
\end{table}

\begin{table*}[t]
\centering

\resizebox{\textwidth}{!}{
\begin{tabular}{l c cc cc c ccc} 
\toprule
\multirow{3}{*}{Method} & \multirow{3}{*}{BlocksWorld} & \multicolumn{5}{c}{TravelPlanner} & \multicolumn{3}{c}{T-Eval}  \\
\cmidrule(lr){3-7} \cmidrule(lr){8-10}
& & \multicolumn{2}{c}{Commonsense} & \multicolumn{2}{c}{Hard} & \multirow{2}{*}{Final} & \multirow{2}{*}{Precision} & \multirow{2}{*}{Recall} & \multirow{2}{*}{F1-score} \\
\cmidrule(lr){3-4} \cmidrule(lr){5-6}

& & Micro & Macro & Micro & Macro & & & & \\
\midrule
DeepSeek-R1-Distill-Llama-8B & 1.4 & 61.2 & 0.0 & 0.0 & 0.0 & 0.0 & 81.8 & 79.3 & 79.4 \\
+SFT        & 12.6 & 78.1 & 16.7 & 29.8 & 16.1 & 3.9  & 87.1 & \underline{86.5} & 86.0  \\
+RFT        & \underline{42.0} & \underline{80.8} & \underline{25.0} & \underline{36.2} & \underline{19.4} & \underline{5.6} & \underline{88.0} & 86.0 & \underline{86.4} \\
+CARL (Ours)       & \textbf{52.6} & \textbf{81.1} & \textbf{32.1} & \textbf{42.9} & \textbf{28.9} & \textbf{11.7} & \textbf{88.4} & \textbf{88.6} & \textbf{87.5} \\

\midrule

Qwen3-8B    & 31.2 & 72.7 & 7.8  & 34.8 & 27.8 & 2.2  & 86.6 & 83.5 & 84.2  \\
+SFT        & 14.2 & 77.4 & 25.0 & 24.3 & 14.4 & 8.3  & 84.0 & 84.3 & 83.5  \\
+RFT        & \underline{73.8} & \underline{96.3} & \underline{74.4} & \underline{65.7} & \underline{48.9} & \underline{45.0} & \underline{88.6} & \underline{88.2} &  \underline{87.3} \\
+CARL (Ours)       & \textbf{77.2} & \textbf{97.3} & \textbf{81.7} & \textbf{73.1} & \textbf{59.4} & \textbf{56.1} & \textbf{89.5} & \textbf{88.5} & \textbf{88.1} \\
\bottomrule
\end{tabular}
}
\caption{Performance comparison with SFT on planning benchmarks. The best and second-best results are \textbf{bold} and \underline{underlined}.}
\label{tab:PC_SFT}
\end{table*}

\subsection{Training Efficiency}
\label{sec:appendix_TE}
To evaluate training efficiency, we measure the average time per training step (in seconds) and planning accuracy across different settings (Qwen3-8B is used here). As shown in Table~\ref{tab:5}, increasing rollouts from 8 to 16 modestly improves RFT but increases about 70\% more computational cost. In contrast, CARL achieves better performance with only a 10\% overhead, striking a better balance between efficiency and effectiveness.  
This advantage is due to CARL’s design: although it generates an additional set of unconstrained queries, these require only a single forward pass for log-probabilities and are excluded from the expensive rollout process.

\section{Implementation Details for GRPO}
\label{sec:appendix_ID_GRPO}
We herein present more details about benchmarks, training, and inference for GRPO.
\paragraph{Benchmarks.}
For BlocksWorld, we employ the official one-shot setting. For TravelPlanner, we adopt the "sole-planning" mode to focus on the LLMs’ planning ability, excluding the influence of information gathering abilities required in the "two-stage" mode. For T-Eval, we use the default setting.

\paragraph{Training.}
For GRPO, we employ the Verl framework. Regarding hyperparameters, we set the policy LLM learning rate to 1e-6 and sample 8 responses per query. Training is conducted on a single node with 8 A100 GPUs, with a total batch size of 64. The maximum response length is set to 8,192 tokens. To optimize GPU memory usage, we enable gradient checkpointing and use Fully Sharded Data Parallel (FSDP) with CPU offloading.

For efficient LLM rollouts, we adopt vllm with a tensor parallel size of 1 and GPU memory utilization ratio of 0.6. The rollout sampling uses a temperature of 0.6 and a top-p value of 1.0. The weighting coefficients $\alpha$ and $\beta$ (see Equation~\ref{eq:objective}) are both set to 0.001.

We train the model for 150 steps on BlocksWorld and T-Eval, and 300 steps on the more challenging TravelPlanner. The task reward is computed using the evaluation script provided by each benchmark. 

\paragraph{Inference.}
During inference, we employ vllm with a sampling temperature of 0.6 and set the maximum response length uniformly to 30,000 tokens to prevent truncation.

\section{Additional Experimental Results}
\label{sec:appendix_MRPB}
We herein present additional experimental results, including comparisons with SFT, prompting, and agent-based methods, generalization performance on other benchmarks, and more results and additional analyses on BlocksWorld.

\subsection{Comparison with SFT}
\label{sec:appendix_CSFT}
Table~\ref{tab:PC_SFT} shows the performance comparison with SFT on planning benchmarks.

\paragraph{Analysis.}
The experimental results demonstrate that RL-based methods are more effective than SFT at improving planning capabilities. We attribute this to two primary reasons. First, the ground-truth data provided by these benchmarks often contains only the final answer without the intermediate reasoning process. Consequently, direct SFT is not conducive to the model's ability to "think" and learn high-quality planning patterns. Second, planning tasks may have multiple feasible solutions that satisfy the given constraints, yet the ground truth typically provides only one. SFT may thus cause the model to memorize this specific ground-truth solution rather than genuinely learning how to plan.

\begin{table}[t]
  \centering
  
  \resizebox{\columnwidth}{!}{

  \begin{tabular}{lcc} 
    \toprule
    Method & BlocksWorld & TravelPlanner \\
    \midrule
    \multicolumn{3}{l}{\textit{Prompting (Base: GPT-4o)}} \\ 
    Direct & 42.4 & 7.8 \\
    Behavioral Cloning & 51.2 & 9.0 \\
    Oracle Feedback & 52.0 & 11.7 \\
    Reference & 51.4 & 8.3 \\
    \midrule
    \multicolumn{3}{l}{\textit{RL Training (Base: Qwen3-8B)}} \\ 
    CARL (Ours) & 77.2 & 56.1 \\
    \bottomrule
  \end{tabular}
  }
  \caption{Performance comparison with prompting-based methods on planning benchmarks. Metrics: accuracy for BlocksWorld and final pass rate for TravelPlanner.}
  \label{tab:PC_PM}
\end{table}

\subsection{Comparison with Prompting-based Methods}
\label{sec:appendix_CPM}
We herein introduce three prompting-based methods for planning performance comparison. Following previous work~\citep{zhao2024expel,fu2024autoguide}, these methods summarize insights from previous attempts to guide the agent:

\begin{itemize}
    \item Behavioral Cloning: The agent is provided with previous failed attempts along with a ground truth plan.
    \item Oracle Feedback: The agent is provided with previous failed attempts accompanied by feedback from a solver or evaluator that explicitly explains the reasons for failure.
    \item Reference: This setting utilizes human-written insights to serve as a ground truth summary of constraints.
\end{itemize}

\paragraph{Analysis.}
We compare these prompting strategies, implemented using the powerful GPT-4o, against our CARL framework trained on the much smaller Qwen3-8B. As presented in Table~\ref{tab:PC_PM}, while prompting methods like oracle feedback improve over the direct baseline, they remain significantly inferior to CARL. Notably, CARL achieves a 77.2\% accuracy on BlocksWorld and a 56.1\% final pass rate on TravelPlanner, far surpassing even the best prompting results (52.0\% and 11.7\%, respectively). This demonstrates that while in-context guidance provides marginal gains, CARL effectively internalizes constraint awareness, allowing a smaller model to outperform larger models relying on extensive prompt engineering.

\begin{table*}[t]
\centering
\begin{tabular}{llcccccc}
\toprule

\multirow{3}{*}{Method} & \multirow{3}{*}{Base Model} & \multirow{3}{*}{External Solver} & \multicolumn{5}{c}{TravelPlanner} \\
\cmidrule(lr){4-8}
& & & \multicolumn{2}{c}{Commonsense} & \multicolumn{2}{c}{Hard} & \multirow{2}{*}{Final} \\
\cmidrule(lr){4-5} \cmidrule(lr){6-7}
& & & Micro & Macro & Micro & Macro & \\
\midrule

LLM-Modulo & \multirow{2}{*}{GPT-4} & \multirow{2}{*}{No} & - & - & - & - & 20.0 \\
Multi-Agent & & & 90.0 & 41.7 & 55.7 & 48.3 & 31.7 \\
\midrule
\multirow{2}{*}{SMT Solver} & GPT-4 & \multirow{2}{*}{Yes} & 95.0 & 95.0 & 95.7 & 98.9 & 93.3 \\
 & Mistral-Large & & 72.0 & 70.6 & 63.3 & 66.7 & 66.7 \\
\midrule
CARL (Ours) & Qwen3-8B & No & 97.3 & 81.7 & 73.1 & 59.4 & 56.1 \\
\bottomrule
\end{tabular}
\caption{Performance comparison with agent-based methods on TravelPlanner.}
\label{tab:6}
\end{table*}

\subsection{Comparison with Agent-based  Methods}
\label{sec:appendix_CAM}
We herein introduce three agent-based methods for planning performance comparison:

\begin{itemize}
\item LLM-Modulo (ICML 2024): A planning agent based on feedback from external constraint critics from \citet{kambhampati2024llms}.
\item Multi-Agent (COLING 2025): A collaborative multi-agent system for planning based on task decomposition from \citet{zhang2024planning}.
\item SMT Solver (NAACL 2025): A planning agent based on results of an external optimization solver from \citet{hao2024large}. 
\end{itemize}

\paragraph{Results.}
We chose the complex real-world planning task TravelPlanner for comparison. Notably, TravelPlanner operates in two modes: "sole-planning" and "two-stage". The former focuses on LLMs’ planning ability, with all relevant travel information provided as a lengthy textual context within the prompt, while the latter requires LLMs to gather information and complete planning.
The aforementioned agent-based methods are specifically designed with tailored prompts for the "two-stage" mode and cannot be directly adapted to the "sole-planning" mode (which we use in the main results). We therefore report their performance under the "two-stage" mode, with results shown in Table~\ref{tab:6}.

\paragraph{Analysis.}
The experimental results demonstrate that while current agent-based methods (e.g., LLM-Modulo and Multi-Agent system) exhibit strong performance in complex planning tasks, their effectiveness typically relies on top models like GPT-4 combined with external tools (e.g., constraint critics) or multi-agent collaboration mechanisms to complete reasoning and planning processes. In contrast, our proposed CARL framework differs fundamentally in design philosophy: it neither depends on top models nor introduces external solvers or multi-agent architectures, but instead achieves end-to-end scalable constraint-aware planning within a single lightweight model.

Notably, using only the Qwen3-8B model – a medium-scale model – CARL still outperforms two of the three agent-based baselines (LLM-Modulo and Multi-Agent). This indicates that by internalizing constraint comprehension within the model's planning process, CARL effectively compensates for its scale limitations while demonstrating robust planning capabilities.

Although the SMT solver-based method currently achieves state-of-the-art performance, its superiority heavily depends on a tightly-coupled GPT-4 + external solver architecture. We observe significant performance degradation when replacing GPT-4 with the equally capable 123B Mistral-Large model, revealing excessive dependency on specific architectures (particularly GPT-4). \textbf{Furthermore, this method requires external solvers to search through vast solution spaces, incurring substantial latency (averaging 245 seconds per query on TravelPlanner) – over 24× slower than CARL's inference efficiency (10s average).} Such high latency and strong dependencies limit its practical scalability and deployment flexibility.

Given the fundamental differences between CARL and existing agent-based methods in system architecture, model dependency, planning workflow, and external tool usage, direct "end-to-end" performance comparisons may lack fairness and interpretability due to inconsistent experimental setups. We therefore exclude direct performance comparisons with these methods in the main results to ensure experimental rigor and comparability.

Nevertheless, it must be emphasized that CARL achieves planning capabilities comparable to or even surpassing those of GPT-4-based agent methods with complex engineering architectures – all without external solvers, top models, or exceeding 8B parameters. This breakthrough highlights CARL's significant potential in empowering medium- and small-scale language models to achieve autonomous, efficient, and compliant planning, while establishing a new technical pathway for developing independent, lightweight, and reliable intelligent agents.

\begin{table*}[t]
\centering

\begin{tabular}{lccccccc|cc}
\toprule
\multirow{2}{*}{Method} & \multicolumn{7}{c|}{\textbf{ALFWorld}} & \multicolumn{2}{c}{\textbf{WebShop}} \\
 & Pick & Look & Clean & Heat & Cool & Pick2 & All & Score & Succ. \\ 
\midrule
Qwen3-8B & 58.6 & 33.3 & {12.9} & {10.5} & 4.3 & {10.0} & {21.9} & 7.1 & 0.8 \\
+RFT & 55.2 & 50.0 & 6.5 & {10.5} & 0.0 & 5.0 & 18.8 & 5.3 & 0.8 \\
+CARL & {58.6} & {66.7} & 9.7 & 0.0 & {4.3} & {10.0} & 21.1 & {8.1} & {1.6} \\ 
\bottomrule
\end{tabular}

\caption{Generalization performance from BlocksWorld to ALFWorld and WebShop.}
\label{tab:GP_AW}
\end{table*}

\begin{table}[t]

\centering

\resizebox{\columnwidth}{!}{
\begin{tabular}{lccccc}
\toprule

\multirow{3}{*}{Method} & \multicolumn{5}{c}{TripCraft} \\
\cmidrule(lr){2-6}
& \multicolumn{2}{c}{Commonsense} & \multicolumn{2}{c}{Hard} & \multirow{2}{*}{Final} \\
\cmidrule(lr){2-3} \cmidrule(lr){4-5}
& Micro & Macro & Micro & Macro & \\
\midrule

Qwen3-8B & 90.8 & 0.0 & 17.3 & 14.8 & 0.0   \\
+RFT    & 90.1 & 1.8 & 23.2 & 21.3 & 0.2 \\
+CARL   & 94.8 & 2.9 & 27.3 & 25.7 & 1.1 \\
\bottomrule
\end{tabular}
}
\caption{Generalization performance from TravelPlanner to TripCraft.}
\label{tab:GP_TC}
\end{table}

\subsection{Generalization Performance on Other Benchmarks}
\label{sec:appendix_GPOB}

We herein introduce three additional benchmarks to verify the generalization ability of CARL. ALFWorld and WebShop serve as \textbf{multi-turn interactive tasks without explicit constraints} to test generalization from BlocksWorld, while TripCraft serves as \textbf{a more complex travel planning task} to test generalization from TravelPlanner.

\begin{itemize}
    \item ALFWorld (ICLR 2021): An embodied environment designed to assess the ability of LLM agents to perform multi-step decision-making~\citep{shridhar2020alfworld}. In each episode, the agent receives a text goal and must accomplish it through multi-turn interaction with the environment. It includes 3,827 task instances across six categories of common household activities.
    \item WebShop (NeurIPS 2022): A complex, web-based interactive environment designed to test LLM agents in realistic online shopping scenarios~\citep{yao2022webshop}. To complete the task, the agent must interact with a simulated HTML-based shopping website to search for, navigate to, and ultimately purchase a suitable item from over 1.1 million products.
    \item TripCraft (ACL 2025): A real-world travel planning benchmark constructed entirely from real data sources to ensure geographical coherence~\citep{chaudhuri2025tripcraft}. Unlike previous datasets reliant on semi-synthetic data, TripCraft integrates complex real-world constraints such as public transit schedules and event availability. This makes it a more challenging and realistic testbed for evaluating the generalization capabilities of planning agents.
\end{itemize}

\paragraph{Settings.}
We evaluate the cross-domain generalization of models trained on specific tasks. First, we take the models trained on BlocksWorld and evaluate them on ALFWorld and WebShop without any fine-tuning. For ALFWorld, we report the success rate (\%) for each of the six subtasks as well as the overall result. For WebShop, we report both the score and the success rate (\%). Second, to test generalization in planning scenarios, we evaluate the models trained on TravelPlanner on the more realistic TripCraft benchmark. We adopt the "w/o Parameter Info" and “3-day” settings for TripCraft and report the micro/macro scores for commonsense and hard constraints, along with the final pass rate.

\paragraph{Analysis.}
Results in Table~\ref{tab:GP_AW} and Table~\ref{tab:GP_TC} reveal two key insights regarding generalization. 
First, enhanced planning capabilities contribute to robust multi-step decision-making. Standard RFT tends to overfit the source domain (BlocksWorld), leading to negative transfer on unseen interactive tasks (e.g., performance drops on ALFWorld compared to the base model). 
In contrast, CARL not only mitigates this degradation on ALFWorld but surprisingly outperforms both RFT and the base model on WebShop. 
This suggests that the constraint-aware optimization in CARL does not merely memorize specific constraints; instead, it refines the model's fundamental reasoning capability. By learning to adhere to constraints, the model becomes more deliberate in its planning, which translates effectively into better performance in complex, multi-turn decision-making environments.

Second, CARL demonstrates superior transferability in complex planning scenarios. On TripCraft, which shares the planning nature of TravelPlanner but imposes significantly harder real-world constraints, CARL achieves a substantial improvement over baselines (e.g., a 25.7 macro score on hard constraints vs. 14.8 for base). This confirms that CARL successfully internalizes a generalized constraint-aware mechanism that remains effective across distinct planning domains.

\subsection{More Results on BlocksWorld}
\label{sec:appendix_MRB}

\begin{table}[t]

  \centering
  \begin{tabular}{lcc}
\toprule
Method & Qwen-7B  & Qwen-14B \\
\midrule
RFT    & 24.6         & 86.0     \\
CARL   & 29.8         & 93.0     \\
\bottomrule
\end{tabular}
  \caption{More results with DeepSeek-R1-Distill models on BlocksWorld.}
  \label{tab:2}
\end{table}

We apply CARL to other DeepSeek-R1-Distill models and compare it against RFT. As shown in Table~\ref{tab:2}, CARL consistently outperforms RFT. These results, combined with the performance of DeepSeek-R1-Distill-Llama-8B in Table~\ref{tab:1}, suggest that CARL can generalize to different model architectures. Notably, CARL achieves performance on par with top models at the 14B scale on BlocksWorld.

\subsection{Additional Analyses on BlocksWorld}
\label{sec:appendix_CR_robustness}

To further validate the practical behavior of CARL, we provide two additional compact analyses: threshold filtering and robustness to noisy task rewards.

\paragraph{Threshold Filtering Ablation.}
We test whether filtering low-divergence samples improves performance. For CA-DPO (1,600 pairs), we filter out the bottom-$k\%$ pairs by KL divergence during pair construction. For CA-GRPO, we apply a hard mask to $R_{\text{CA}}$ for the bottom-$k\%$ samples in each batch. Results are shown in Table~\ref{tab:appendix_threshold_filtering}.
\begin{table}[t]
  \centering
  \begin{tabular}{lcc}
\toprule
Threshold & CA-DPO & CA-GRPO \\
\midrule
0\%   & 53.8 & 77.2 \\
0.5\% & 53.4 & 77.2 \\
1.0\% & 53.6 & 76.8 \\
2.0\% & 53.0 & 77.0 \\
5.0\% & 52.6 & 76.6 \\
\bottomrule
\end{tabular}
  \caption{Effect of threshold filtering on BlocksWorld.}
  \label{tab:appendix_threshold_filtering}
\end{table}
Small thresholds do not provide consistent gains, while aggressive filtering degrades performance, indicating that CARL already self-regulates effectively without hard filtering.

\paragraph{Noise Injection Robustness.}
To simulate imperfect reward signals, we randomly flip the binary task reward during GRPO training on BlocksWorld (Qwen3-8B). Results are shown in Table~\ref{tab:appendix_noise}.
\begin{table}[t]
  \centering
  \begin{tabular}{lccc}
\toprule
Noise Ratio & RFT & CARL & Difference \\
\midrule
0\%  & 73.8 & 77.2 & +3.4 \\
5\%  & 69.6 & 76.2 & +6.6 \\
10\% & 66.4 & 74.4 & +8.0 \\
15\% & 59.8 & 67.6 & +7.8 \\
20\% & 50.2 & 56.4 & +6.2 \\
\bottomrule
\end{tabular}
  \caption{Noise injection robustness on BlocksWorld.}
  \label{tab:appendix_noise}
\end{table}
CARL maintains consistent advantages under all tested noise levels, suggesting stronger robustness when external reward signals are imperfect.

\section{Implementation Details for DPO}
\label{sec:appendix_ID_DPO}

We herein provide more details about preference pair construction and training for DPO.

\begin{table}[t]
  \centering
  \resizebox{\columnwidth}{!}{
  \begin{tabular}{l|c|c}
\toprule
Sampling Strategy  & Chosen Samples & Rejected Samples \\
\midrule
Naive-DPO        & $  y^{(i)} \in \mathcal{Y}_{\text{valid}}(x^{(i)})  $ & $  y^{(i)} \notin \mathcal{Y}_{\text{valid}}(x^{(i)})  $ \\
CA-DPO             & $  y^{(i)} \in \mathcal{Y}_{\text{valid}}(x^{(i)})  $ & $  y^{(i)}_{\setminus\mathcal{C}}  $ \\
\bottomrule
\end{tabular}
}
  \caption{The sampling strategies used in Naive-DPO and CA-DPO. In CA-DPO, the rejected samples are outputs generated under conditions where constraints are absent, represented as $  y^{(i)}_{\setminus\mathcal{C}}  $. This design choice aims to maximize the divergence between $  y^{(i)}_{\setminus\mathcal{C}}  $ and the valid constrained output $  y^{(i)}  $. By doing so, the model is better guided to understand and incorporate the crucial role played by constraints in its learning process.}
  \label{tab:7}
\end{table}

\paragraph{Preference Pair Construction.}
For Naive-DPO, we input constrained queries and construct preference pairs based on the correctness of the responses. For CA-DPO, we input both constrained and unconstrained queries. The correct responses generated for constrained queries are selected as positive samples, while the responses generated for unconstrained queries are randomly chosen as negative samples, which are then utilized to construct preference pairs.
Our sampling strategies for preference pair construction can be summarized as shown in Table~\ref{tab:7}.

\paragraph{Training.}
For DPO, we employ the LLaMA-Factory framework. We consistently train the model for 3 epochs with a batch size of 64, a learning rate of 2.0e-6, a preference coefficient $\beta$ of 0.1, and a warmup ratio of 0.05 on BlocksWorld.

\begin{figure}[t]
\centering
  \includegraphics[width=\columnwidth]{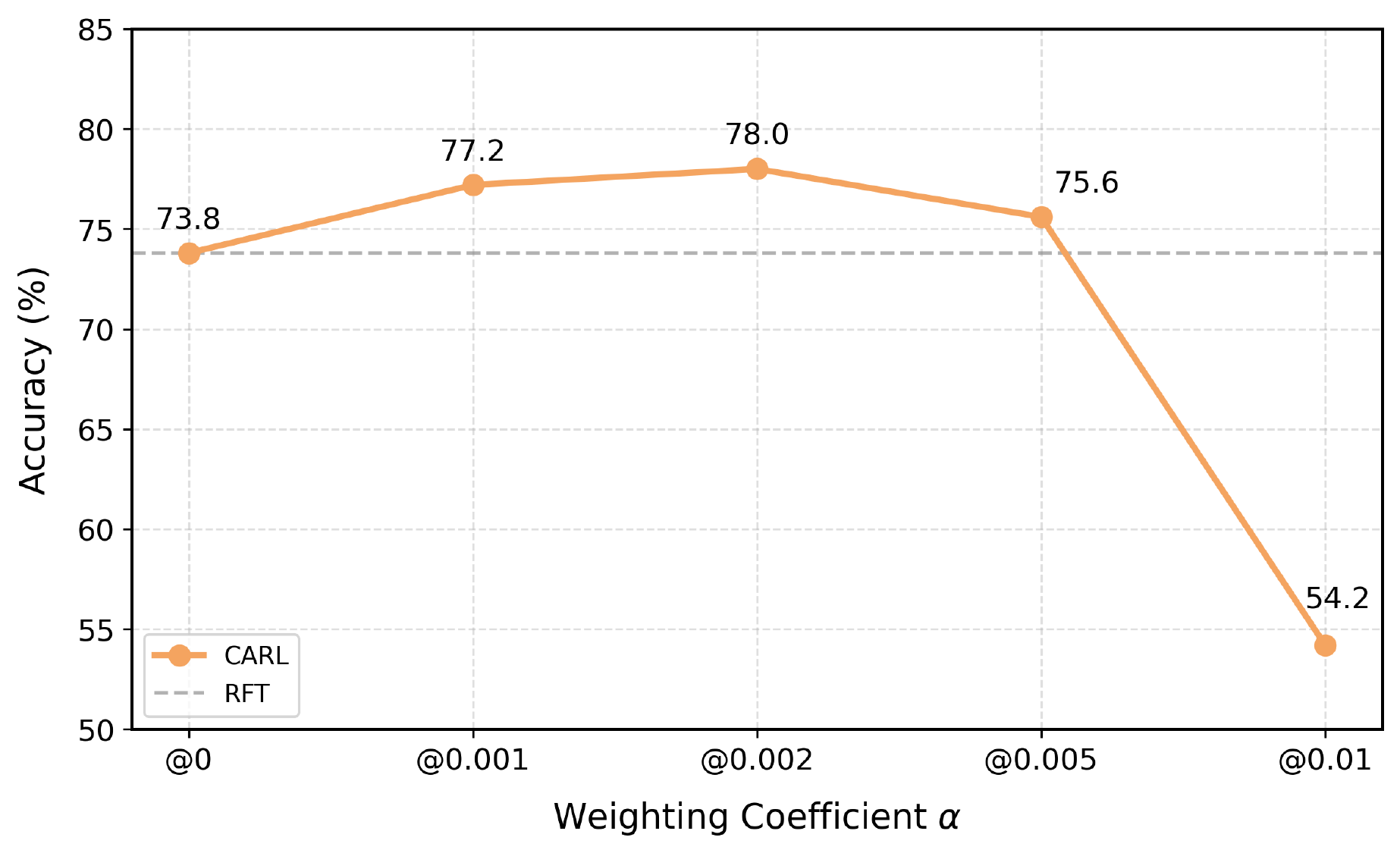}
  
  \caption{Sensitivity analysis of constraint-aware reward weight $\alpha$ on BlocksWorld. The KL penalty coefficient is fixed at $\beta=0.001$.}
  \label{fig:fig_sacarw}
\end{figure}

\section{Sensitivity Analysis of Constraint-Aware Reward Weight}
\label{sec:appendix_sacarw}

We herein investigate the sensitivity of the CARL framework to the weighting coefficient $\alpha$ of the constraint-aware reward.

\paragraph{Settings.}
We conduct the ablation study on the BlocksWorld benchmark using the Qwen3-8B model. While keeping the KL penalty coefficient fixed at $\beta=0.001$, we vary $\alpha$ across the set $\{0, 0.001, 0.002, 0.005, 0.01\}$. Note that $\alpha=0$ is equivalent to the standard RFT baseline. We report the accuracy for each setting to analyze the impact of $\alpha$ on model performance.

\begin{figure*}[t]
\centering
    \includegraphics[width=\linewidth]{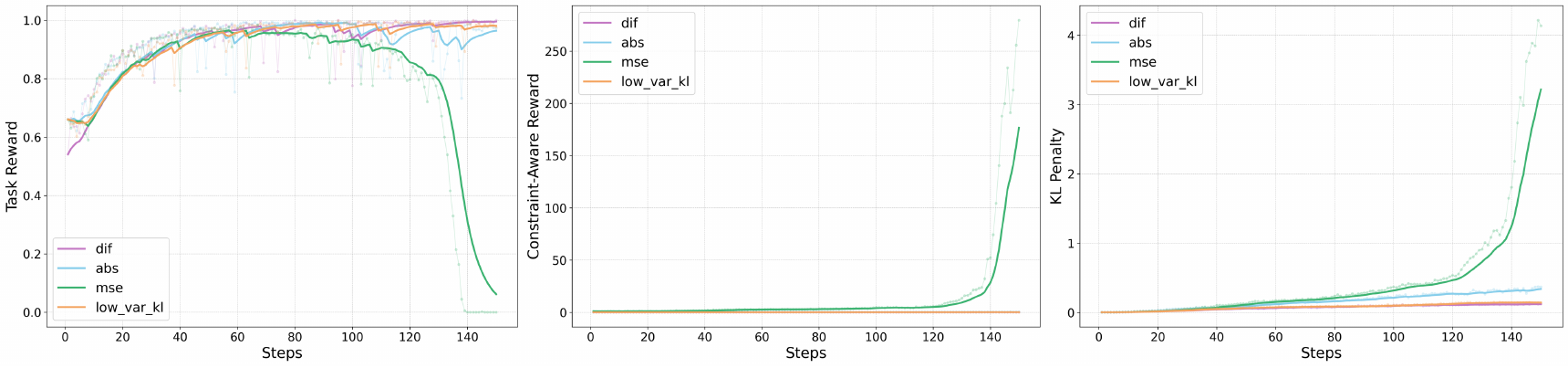}
    \caption{Comparison of the training dynamics on BlocksWorld. Solid lines indicate exponential moving averages of the data.}
    \label{fig:fig7}
\end{figure*}

\paragraph{Analysis.}
The results, illustrated in Figure~\ref{fig:fig_sacarw}, demonstrate that CARL is highly robust to the selection of $\alpha$. The method consistently outperforms the RFT baseline across a broad range of values, from $\alpha=0.001$ to $\alpha=0.005$. Notably, while our main experiments adopted a conservative setting of $\alpha=0.001$, the performance peaks at $\alpha=0.002$, and even at $\alpha=0.005$—a five-fold increase—the model maintains a competitive accuracy of $75.6\%$. This indicates that CARL does not require precise hyperparameter tuning to yield improvements. However, performance degrades far below the baseline at $\alpha=0.01$, suggesting that an excessively strong constraint signal overwhelms the primary task objectives.

\section{Impact of Reward Calculation Strategy on Training Dynamics}
\label{sec:appendix_IRCSTD}

The results from the ablation study, presented in Table~\ref{tab:AS_RCS}, clearly show that the choice of reward calculation metric has a profound impact on training outcomes. While the dif, abs, and low\_var\_kl strategies all yield high and comparable final performance (76.6, 76.0, and 77.2, respectively), the mse strategy leads to a complete training collapse, with the model achieving a near-zero score of 0.2.

The training dynamics, illustrated in Figure~\ref{fig:fig7}, reveal the underlying reasons for this disparity. The mse strategy (green line) initially shows a promising increase in task reward, but it abruptly collapses after approximately 120 steps. This collapse is directly correlated with an explosive, uncontrolled spike in both the constraint-aware reward and the KL penalty, indicating severe training instability. We hypothesize that the quadratic nature of mse creates a volatile reward signal, excessively penalizing larger deviations and leading to policy divergence.

In contrast, dif, abs, and low\_var\_kl all maintain a stable and high task reward throughout training. Their constraint-aware rewards remain close to zero, providing a consistent and stable learning signal. Notably, low\_var\_kl and dif also maintain the lowest KL penalty, suggesting they guide the policy towards improvement more efficiently without deviating drastically from the reference model. This combination of high final performance and superior training stability confirms that low\_var\_kl is the most robust and effective reward calculation strategy for our framework.

\section{Details for Attribution Analysis}
\label{sec:appendix_DAA}
\paragraph{Setup.}
In this paper, we adopt Feature Ablation (implemented via \texttt{FeatureAblation} in Captum~\citep{kokhlikyan2020captum}) as our analysis strategy for testing the inner
workings of LLMs when planning. For TravelPlanner, we conduct analysis using the entire validation set (180 samples). For BlocksWorld, we randomly select 200 samples from the test set due to the high computational cost.

\paragraph{Case Study.}
We present several cases used in our attribution analysis. The model after RFT fails in planning due to its insufficient sensitivity to constraints, whereas the model after CARL succeeds in planning by adequately attending to the constraints. Due to the length of the full response, we present only the portion following the "\texttt{</think>}" identifier.

The case on BlocksWorld demonstrates that the model after RFT executes a pick-up action immediately following an unstack operation, violating the fundamental constraint that prohibits pick-up or unstack operations while holding a block. The related constraints are explicitly highlighted with underlines within the case.

The case on TravelPlanner demonstrates that the model after CARL proactively considers whether daily accommodation and meal arrangements align with user preferences, thus producing plans that fully satisfy all constraints. This focus on preferences is partially explicitly highlighted with double underscores within the case.

\clearpage
\onecolumn

\lstset{
    numbers=none, 
    backgroundcolor=\color[RGB]{245,245,245},
    breaklines=true,
    breakindent=0pt,
    basicstyle=\ttfamily\small,
    frame=trbl,
    frameround = tttt,
    escapeinside={(*@}{@*)} 
}\begin{lstlisting}
(*@\textbf{[BlocksWorld]}@*)

(*@\textbf{Query:}@*)
I am playing with a set of blocks where I need to arrange the blocks into stacks. Here are the actions I can do

Pick up a block
Unstack a block from on top of another block
Put down a block
Stack a block on top of another block

I have the following restrictions on my actions:
I can only pick up or unstack one block at a time.
(*@\underline{I can only pick up or unstack a block if my hand is empty.}@*)
I can only pick up a block if the block is on the table and the block is clear. A block is clear if the block has no other blocks on top of it and if the block is not picked up.
I can only unstack a block from on top of another block if the block I am unstacking was really on top of the other block.
I can only unstack a block from on top of another block if the block I am unstacking is clear.
(*@\underline{Once I pick up or unstack a block, I am holding the block.}@*)
I can only put down a block that I am holding.
I can only stack a block on top of another block if I am holding the block being stacked.
I can only stack a block on top of another block if the block onto which I am stacking the block is clear.
Once I put down or stack a block, my hand becomes empty.
Once you stack a block on top of a second block, the second block is no longer clear.

[STATEMENT]
As initial conditions I have that, the blue block is clear, the hand is empty, the red block is on top of the orange block, the blue block is on top of the yellow block, the yellow block is on top of the red block and the orange block is on the table.
My goal is to have that the blue block is on top of the red block and the yellow block is on top of the orange block.

My plan is as follows:

[PLAN]
unstack the blue block from on top of the yellow block
put down the blue block
unstack the yellow block from on top of the red block
stack the yellow block on top of the blue block
unstack the red block from on top of the orange block
put down the red block
unstack the yellow block from on top of the blue block
stack the yellow block on top of the orange block
pick up the blue block
stack the blue block on top of the red block
[PLAN END]

[STATEMENT]
As initial conditions I have that, the red block is clear, the orange block is clear, the hand is empty, the red block is on top of the yellow block, the orange block is on top of the blue block, the blue block is on the table and the yellow block is on the table.
My goal is to have that the blue block is on top of the yellow block and the yellow block is on top of the red block.

My plan is as follows:

[PLAN]

(*@\textbf{RFT Response:}@*)
The user's plan contains errors due to misalignment with the initial conditions. Here's the corrected plan:

Correct Plan:

1. Unstack the red block from the yellow block  
   - Red is on top of yellow; red is clear (no blocks above it).  
   - Unstack red and place it on the table.  

2. Pick up the yellow block  
   - Yellow is now on the table (after red is removed) and is clear.  

3. Stack the yellow block on the red block  
   - Red is on the table (clear), so yellow can be stacked on it.  
   - Result: Red (*@{$\rightarrow$}@*) Yellow.  

4. Unstack the orange block from the blue block  
   - Orange is on top of blue; orange is clear (no blocks above it).  
   - Unstack orange and place it on the table.  

5. Pick up the blue block  
   - Blue is now on the table (after orange is removed) and is clear.  

6. Stack the blue block on the yellow block  
   - Yellow is on top of red and is clear (no blocks above it).  
   - Result: Red (*@{$\rightarrow$}@*) Yellow (*@{$\rightarrow$}@*) Blue.  

Final Arrangement:  
- Red block is on the table.  
- Yellow block is on top of red.  
- Blue block is on top of yellow.  

This sequence adheres to all constraints and achieves the goal. The user's original plan incorrectly assumed blue was on yellow initially, which was not the case.

(*@\textbf{CARL Response:}@*)
To achieve the goal of having blue on top of yellow and yellow on top of red, given the initial configuration:

- Red is on top of yellow
- Orange is on top of blue
- Blue and yellow are on the table
- Red and orange are clear

We need to rearrange the blocks while adhering to the rules of block manipulation. Here's the step-by-step plan:

---

Plan

1. Unstack the red block from on top of the yellow block  
   - Red is clear (initial condition), and it is on top of yellow.  
   - After unstacking, red is held in the hand.  
   - Yellow is now clear (since red is removed from above it).

2. Put down the red block  
   - Hand becomes empty.  
   - Red is now on the table.

3. Pick up the yellow block  
   - Yellow is now clear (no blocks on top).  
   - Hand is empty, so this is allowed.

4. Stack the yellow block on top of the red block  
   - You are holding yellow, and red is clear (before stacking).  
   - After stacking, red is no longer clear (rule: stacking makes the base block non-clear).  
   - Hand becomes empty.

5. Unstack the orange block from on top of the blue block  
   - Orange is clear (initial condition), and it is on top of blue.  
   - After unstacking, orange is held in the hand.  
   - Blue is now clear (since orange is removed from above it).

6. Put down the orange block  
   - Hand becomes empty.  
   - Orange is now on the table.

7. Pick up the blue block  
   - Blue is now clear (no blocks on top).  
   - Hand is empty, so this is allowed.

8. Stack the blue block on top of the yellow block  
   - You are holding blue, and yellow is clear (no blocks on top).  
   - After stacking, yellow is no longer clear (rule: stacking makes the base block non-clear).  
   - Hand becomes empty.

---

Final Arrangement

- Red block is on the table.  
- Yellow block is on top of red.  
- Blue block is on top of yellow.  

This satisfies the goal: blue is on top of yellow, and yellow is on top of red.

---

Answer
```
[PLAN]
unstack the red block from on top of the yellow block
put down the red block
pick up the yellow block
stack the yellow block on top of the red block
unstack the orange block from on top of the blue block
put down the orange block
pick up the blue block
stack the blue block on top of the yellow block
[PLAN END]
```
\end{lstlisting}

\newpage

\lstset{
    numbers=none, 
    backgroundcolor=\color[RGB]{245,245,245},
    breaklines=true,
    breakindent=0pt,
    basicstyle=\ttfamily\small,
    frame=trbl,
    frameround = tttt,
    escapeinside={(*@}{@*)} 
}\begin{lstlisting}
(*@\textbf{[TravelPlanner]}@*)

(*@\textbf{Query:}@*)
I'm looking for a 7-day travel itinerary for 2 people, starting from Reno and heading to Texas, specifically visiting 3 different cities. The travel dates are from March 7th to March 13th, 2022, with a set budget of $4,300. We require accommodations that adhere to house rules regarding visitors and should ideally be entire rooms. For food, we would love to try a variety of cuisines, including Chinese, French, American, and Mediterranean.

(*@\textbf{RFT Response:}@*)
Travel Plan:  
Day 1:  
Current City: from Reno to Abilene  
Transportation: Self-driving, from Reno to Abilene, Duration: 22 hours 27 mins, Distance: 2,412 km, Cost: 120  
Breakfast: Thai Garden, Abilene  
Attraction: The Grace Museum, Abilene  
Lunch: Crispy Crust, Abilene  
Dinner: Mx Corn, Abilene  
Accommodation: Apartment minutes from manhattan, Abilene  

Day 2:  
Current City: Abilene  
Transportation: -  
Breakfast: Cakes Degree, Abilene  
Attraction: Frontier Texas!, Abilene; Historic Fort Phantom Hill, Abilene  
Lunch: LPK Waterfront, Abilene  
Dinner: Pawan Foods, Abilene  
Accommodation: Apartment minutes from manhattan, Abilene  

Day 3:  
Current City: from Abilene to Amarillo  
Transportation: Self-driving, from Abilene to Amarillo, Duration: 4 hours 10 mins, Distance: 446 km, Cost: 22  
Breakfast: Wood Box Cafe, Amarillo  
Attraction: Cadillac Ranch, Amarillo; Amarillo Botanical Gardens, Amarillo  
Lunch: Anand Restaurant, Amarillo  
Dinner: Thalaivar, Amarillo  
Accommodation: Trendy Brooklyn Room - 20mins from Manhattan, Amarillo  

Day 4:  
Current City: Amarillo  
Transportation: -  
Breakfast: The Cinnamon Kitchen, Amarillo  
Attraction: Amarillo Zoo, Amarillo; Don Harrington Discovery Center, Amarillo  
Lunch: Punjabi Chaap Corner, Amarillo  
Dinner: Cafe Coffee Day, Amarillo  
Accommodation: Trendy Brooklyn Room - 20mins from Manhattan, Amarillo  

Day 5:  
Current City: from Amarillo to Lubbock  
Transportation: Self-driving, from Amarillo to Lubbock, Duration: 1 hour 47 mins, Distance: 197 km, Cost: 9  
Breakfast: Grand Barbeque Buffet Restaurant, Lubbock  
Attraction: Buddy Holly Center, Lubbock; National Ranching Heritage Center, Lubbock  
Lunch: Kapoor's Sanjha Chulha, Lubbock  
Dinner: San Carlo, Lubbock  
Accommodation: Gorgeous Spacious Room in Clinton Hill, Lubbock  

Day 6:  
Current City: Lubbock  
Transportation: -  
Breakfast: Assam Tea Corner, Lubbock  
Attraction: American Windmill Museum, Lubbock; Lubbock Lake National Historic Landmark, Lubbock  
Lunch: Domino's Pizza, Lubbock  
Dinner: Platinum Lounge, Lubbock  
Accommodation: Gorgeous Spacious Room in Clinton Hill, Lubbock  

Day 7:  
Current City: from Lubbock to Reno  
Transportation: Self-driving, from Lubbock to Reno, Duration: 20 hours 3 mins, Distance: 2,145 km, Cost: 107  
Breakfast: -  
Attraction: -  
Lunch: -  
Dinner: -  
Accommodation: -

(*@\textbf{CARL Response:}@*)
Travel Plan:  
Day 1:  
Current City: from Reno to Abilene  
Transportation: Self-driving, from Reno to Abilene, duration: 22 hours 27 mins, distance: 2,412 km, cost: 120  
Breakfast: -  
Attraction: -  
Lunch: -  
Dinner: Thai Garden, Abilene (*@\underline{\underline{(Average Cost: \$13, Cuisines: Cafe, American, Desserts)}}@*)  
Accommodation: NYC Studio near Central Park and the Hudson River, Abilene (price: $304, room type: Entire home/apt, house rules: No pets, minimum nights: 2)  

Day 2:  
Current City: Abilene  
Transportation: -  
Breakfast: Crispy Crust, Abilene (Average Cost: $54, Cuisines: Desserts, Bakery, Cafe, American, Seafood)  
Attraction: The Grace Museum, Abilene  
Lunch: Mx Corn, Abilene (*@\underline{\underline{(Average Cost: \$62, Cuisines: Tea, Cafe, Italian)}}@*)  
Dinner: The Grand Trunk Road, Abilene (Average Cost: $80, Cuisines: Tea, Pizza, Bakery, BBQ, Chinese, Mediterranean)  
Accommodation: NYC Studio near Central Park and the Hudson River, Abilene  

Day 3:  
Current City: from Abilene to Amarillo  
Transportation: Self-driving, from Abilene to Amarillo, duration: 4 hours 10 mins, distance: 446 km, cost: 22  
Breakfast: -  
Attraction: -  
Lunch: -  
Dinner: Wood Box Cafe, Amarillo (Average Cost: $11, Cuisines: Desserts, French, Fast Food, Cafe, American)  
Accommodation: Spacious retreat, Amarillo (price: $528, room type: Entire home/apt, house rules: No smoking, minimum nights: 2)  

Day 4:  
Current City: Amarillo  
Transportation: -  
Breakfast: Komachi, Amarillo (Average Cost: $71, Cuisines: Desserts, Tea, Fast Food, Chinese, Mediterranean)  
Attraction: Cadillac Ranch, Amarillo  
Lunch: Anand Restaurant, Amarillo (Average Cost: $53, Cuisines: Fast Food, Seafood, Desserts, Italian)  
Dinner: Shahi Food Corner, Amarillo (Average Cost: $90, Cuisines: French, Pizza, Bakery, Mediterranean)  
Accommodation: Spacious retreat, Amarillo  

Day 5:  
Current City: from Amarillo to Lubbock  
Transportation: Self-driving, from Amarillo to Lubbock, duration: 1 hour 47 mins, distance: 197 km, cost: 9  
Breakfast: -  
Attraction: -  
Lunch: -  
Dinner: Annapurna Bhandar, Lubbock (Average Cost: $78, Cuisines: French, Chinese, Desserts, Fast Food)  
Accommodation: Cozy Clean Small Apartment, Lubbock (price: $313, room type: Entire home/apt, house rules: No children under 10 & No parties, minimum nights: 2)  

Day 6:  
Current City: Lubbock  
Transportation: -  
Breakfast: Paris 6 Classique, Lubbock (Average Cost: $91, Cuisines: Desserts, Cafe, Pizza, Mexican, BBQ, Chinese, Seafood)  
Attraction: Buddy Holly Center, Lubbock  
Lunch: Kapoor's Sanjha Chulha, Lubbock (Average Cost: $21, Cuisines: Tea, Cafe, Indian, American)  
Dinner: Spezia Deliveries, Lubbock (Average Cost: $64, Cuisines: Desserts, Italian, Bakery, Cafe, Mediterranean, Seafood)  
Accommodation: Cozy Clean Small Apartment, Lubbock  

Day 7:  
Current City: from Lubbock to Reno  
Transportation: Self-driving, from Lubbock to Reno, duration: 20 hours 3 mins, distance: 2,145 km, cost: 107  
Breakfast: -  
Attraction: -  
Lunch: -  
Dinner: -  
Accommodation: -

\end{lstlisting}

\end{document}